\def\year{2020}\relax
\documentclass[letterpaper]{article} 
\usepackage{aaai20}  
\usepackage{times}  
\usepackage{helvet} 
\usepackage{courier}  
\usepackage[hyphens]{url}  
\usepackage{graphicx} 
\usepackage{graphicx}  
\usepackage{dblfloatfix} 

\usepackage[utf8]{inputenc} 
\usepackage{url}            
\usepackage{booktabs}       
\usepackage{amsfonts}       
\usepackage{nicefrac}       
\usepackage{microtype}      
\usepackage[mathscr]{euscript}
\usepackage{graphicx}
\usepackage{subfigure}
\usepackage{widetext}
\usepackage{amsmath}
\usepackage{bm}
\usepackage{cleveref}
\newcommand{\citet}[1]{\citeauthor{#1}~\shortcite{#1}}
\newcommand{\citep}{\cite}

\usepackage{amsmath}
\usepackage{natbib}
\usepackage{algorithm}
\usepackage{tikz}
\usepackage[noend]{algpseudocode}

\usepackage[authormarkuptext=name,addedmarkup=bf,authormarkupposition=left]{changes}
\usetikzlibrary{arrows,decorations.pathmorphing,backgrounds,positioning,fit,matrix,calc,shapes.geometric,shapes.misc, positioning,decorations.pathreplacing}
\tikzset{
	multiplexer/.style={
		draw,
		trapezium,
		shape border rotate=270,
		minimum size=2pt,
	}  
}
\title{On the Role of Weight Sharing During Deep Option Learning} 
\author{Matthew Riemer\textsuperscript{\rm 1}, Ignacio Cases\textsuperscript{\rm 2}, Clemens Rosenbaum\textsuperscript{\rm 3}, Miao Liu\textsuperscript{\rm 1}, and Gerald Tesauro\textsuperscript{\rm 1}\\
\textsuperscript{\rm 1} IBM Research, Yorktown Heights, NY\\ 
\textsuperscript{\rm 2} Linguistics Department and Stanford NLP Group, AI Lab, Stanford University \\ 
\textsuperscript{\rm 3} College of Information and Computer Sciences, University of Massachusetts Amherst
}
 
\begin{document}

\maketitle

\begin{abstract}
The options framework is a popular approach for building temporally extended actions in reinforcement learning. In particular, the option-critic architecture provides general purpose policy gradient theorems for learning actions from scratch that are extended in time. However, past work makes the key assumption that each of the components of option-critic has independent parameters. In this work we note that while this key assumption of the policy gradient theorems of option-critic holds in the tabular case, it is always violated in practice for the deep function approximation setting. We thus reconsider this assumption and consider more general extensions of option-critic and hierarchical option-critic training that optimize for the full architecture with each update. It turns out that not assuming parameter independence challenges a belief in prior work that training the policy over options can be disentangled from the dynamics of the underlying options. In fact, learning can be sped up by focusing the policy over options on states where options are actually likely to terminate. We put our new algorithms to the test in application to sample efficient learning of Atari games, and demonstrate significantly improved stability and faster convergence when learning long options. 
\end{abstract}

\section{Introduction}

Developing systems that can autonomously create temporal abstractions is a major problem in scaling deep reinforcement learning (RL). \textit{Options} \citep{Options,PrecupThesis} provide a general purpose framework for defining temporally abstract courses of action for learning and planning in RL. This is a very promising direction with the potential to allow for more coherent exploration and improved long term credit assignment by effectively pruning the number of decision nodes. The popular option-critic \cite{OC} learning framework blurs the line between option discovery and option learning. These approaches have achieved success when applied to Q-learning on Atari \citep{OC}, but also with continuous action spaces \citep{oc_continuous} and asynchronous parallelization \citep{Deliberation}. Additionally, this framework was recently extended to the hierarchical option-critic framework \citep{HOC}, which allows networks to learn an arbitrary depth hierarchy of high level (longer) options and low level (shorter) options. 

\begin{figure*}[!t]
    \centering
    \includegraphics[width=0.9\textwidth]{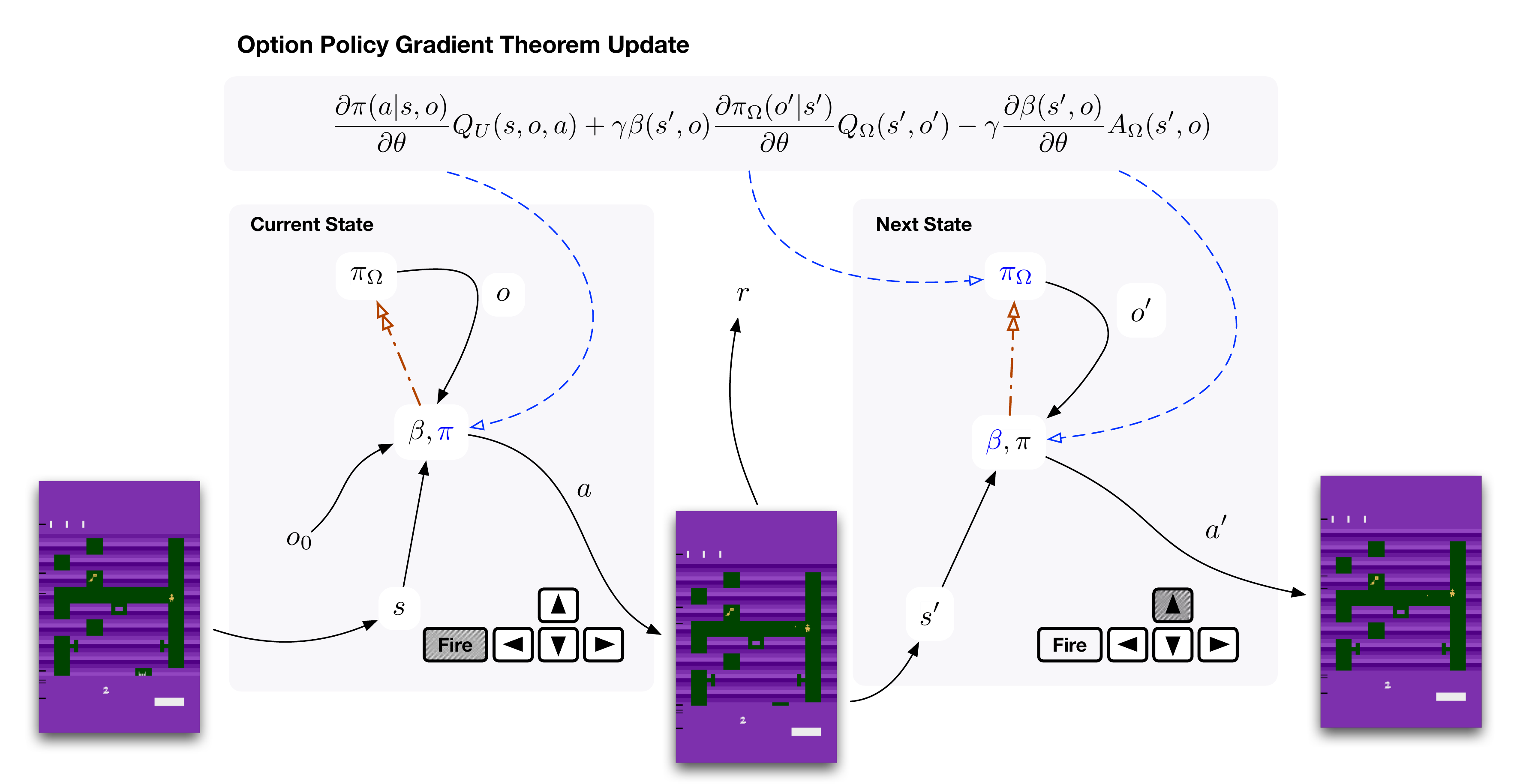}
    \caption{We illustrate the training and execution of an option-critic architecture trained with the generalized option-critic policy gradient update rule on the game Tutankham. Blue lines represent gradients and red lines represent option termination.}
    \label{figure:state_space}
\end{figure*}


With the recent successes of Deep Learning based function approximation applied to RL, a major point of interest has become the derivation of theoretically justified policy gradient theorems. Policy gradient theorems (originally developed for learning with primitive actions by \citet{Sutton2000}) are critical to the success of Deep RL as they define the gradient steps that update the parameters of deep neural networks to maximize the expected reward with gradient descent style learning rules. The option-critic framework \citep{OC} has garnered significant interest largely because it defined the first policy gradient theorems that can be used to update a neural network architecture that is endowed with options that are learned from scratch along with the network. The high level role of options as a temporally extended form of actions is clear. However, how these abstract actions should be composed with respect to the parameters of a neural network is far less clear and has not been closely studied. In fact, the default architecture framework for learning options \citep{OC,HOC} functions in a setting where the underlying assumptions of existing algorithms for optimizing these architectures do not hold. 

Past work on deriving policy gradient theorems for option models \citep{OC} has assumed that the parameters of the option policies $\bm{\pi}$, the policy over options $\bm{\pi_\Omega}$ and the termination functions $\bm{\beta}$ are independent of each other (i.e. $\theta_{\bm{\pi}} \cap \theta_{\bm{\pi_\Omega}} \cap \theta_{\bm{\beta}} = \emptyset$). This assumption gets even more strict for work on hierarchical option models where policies and termination functions are assumed to have independent parameters at each level of the hierarchy as well \citep{HOC}. In practice, while this may hold for tabular problems, this has never been true in application to deep neural networks. In contrast, most parameters have been totally shared using a shared feature extraction network common across all model components and a private output layer for each component. 
This setup closely follows conventions from multi-task learning \citep{Caruana97} with neural networks in the supervised setting. Nonetheless, the policy gradient theorems used to optimize these architectures have not actually been valid. 

In this work, we seek to provide a remedy to this miss-match between the architectures used for deep option learning and the policy gradient theorems used to provide their learning rule. We achieve this by assuming in our derivations that all parameters are shared throughout all components of our model rather than assuming no sharing. This assumption is more general because even if a model component does not use certain parameters, it is valid to optimize for all parameters across all components because each component will only be influenced by the parameters that it uses. To be more concrete, if we assume some global set of shared parameters $\bm{\theta}$ such that $\theta_{\bm{\pi}} \in \bm{\theta}$, $\theta_{\bm{\pi_\Omega}} \in \bm{\theta}$, and $\theta_{\bm{\beta}} \in \bm{\theta}$, we can say for example that the set of parameters $\theta_{\bm{\pi*}} = \bm{\theta} - \theta_{\bm{\pi}}$ induces a zero gradient for $\bm{\pi}$ i.e. $\frac{\partial \bm{\pi}}{\partial \theta_{\bm{\pi*}}}=0$. Past work \citep{OC,HOC} has typically derived separate learning rules for each policy and termination function with respect to their own parameters. By assuming instead to have one shared set of parameters, we arrive at a single learning rule that optimizes for the entire system with respect to its parameters $\bm{\theta}$. We will thus call this 
architecture level learning rule the \textit{option-critic policy gradient} (OCPG) or more generally the \textit{hierarchical option-critic policy gradient} (HOCPG) when modeling deep hierarchies of options. 

Our paper aims to shed light on the disconnect between theory and practice in deep option-critic learning and motivates the benefits of a corrected update rule for efficient learning of long options. Our experiments in challenging RL environments with high dimensional state spaces such as Atari demonstrate the benefits of OCPG and HOCPG 
when using typical strategies for weight sharing across option model components. Additionally, this result is a critical first step towards developing more sophisticated weight sharing schemes across deep option models. It is extremely important that option models allow for more flexibility in this regard as the current methodology \citep{HOC} cannot appropriately handle, for example, the case when the same low level option is called within different high level options. 

\section{Background and Notation}

A Markov Decision Process (MDP) is defined with a set of states $\mathcal{S}$, a set of actions $\mathcal{A}$, a transition function $\mathcal{P}: \mathcal{S} \times \mathcal{A} \rightarrow ( \mathcal{S} \rightarrow [0,1])$ and a reward function $r: \mathcal{S} \times \mathcal{A} \rightarrow \mathbb{R}$. Following standard practice \citep{OC}, we  develop  our  ideas  assuming  discrete  state  and  action sets, while our results extend to continuous spaces using usual measure-theoretic assumptions. A policy is defined as a probability distribution over actions conditioned on states, $\pi: \mathcal{S} \times \mathcal{A} \rightarrow [0,1]$. The value function of a policy $\pi$ is the expected return $V_\pi(s) = \mathbb{E}_\pi[\sum_{t=0}^{\infty} \gamma^t r_{t+1}|s_0=s]$ with an action-value function of $Q_\pi(s,a) = \mathbb{E}_\pi[\sum_{t=0}^{\infty} \gamma^t r_{t+1}|s_0=s,a_0=a]$ where $\gamma \in [0,1)$ is the \emph{discount factor}.

\textbf{Policy gradient methods} improve a  policy  by  performing  gradient  ascent over a family  of parameterized  stochastic  policies, $\pi_\theta$. The policy gradient  theorem  \citep{Sutton2000}  provides an expression to compute the  gradient  of the discounted  reward  objective  with  respect  to $\theta$ and a designated  starting  state $s_0$ in a straightforward expression. 
The theorem defines the gradient update as $\sum_s \mu_{\pi_\theta}(s|s_0) \sum_a \frac{\partial \pi_\theta(a|s)}{\partial \theta} Q_{\pi_\theta}(s,a)$. Here $\mu_{\pi_\theta}(s|s_0) = \sum_{t=0}^{\infty} \gamma^t P(s_t=s|s_0)$ is defined as the  discounted weighting of the states along the trajectories starting from initial state $s_0$.

\textbf{The options framework} \citep{Options,PrecupThesis} formalizes temporally  extended  actions in RL.  A Markovian option $o \in \Omega$ is  a  triple $(I_o,\pi_o,\beta_o)$ where $I_o \subseteq S$ represents an initiation set, $\pi_o$ represents an intra-option policy,  and
$\beta_o: \mathcal{S} \rightarrow [0,1]$ represents  a  termination  function.  Many algorithms (such as option-critic) assume that all options are available everywhere, removing the need to explicitly model $I_o$. MDPs with options become SMDPs \citep{markov1994} with an optimal value function over options $V^*_\Omega(s)$ and option-value function $Q^*_\Omega(s,o)$.  

\textbf{The option-critic architecture} \citep{OC} leverages a call-and-return option execution model where an agent picks option $o$ according to its policy over options $\pi_\Omega(o|s)$, then follows the intra-option policy $\pi(a|s,o)$ until termination (as determined by $\beta(s,o)$). Termination then triggers a repetition of this procedure. Let $\pi_{\theta_\pi}(a|s,o)$ denote the intra-option policy of option $o$ parametrized by $\theta_\pi$ and $\beta_{\theta_\beta}(s,o)$ the termination function of $o$ parameterized by $\theta_\beta$. 
The option-value function is then defined as:

\vspace{-2mm}
\begin{equation} \label{QOmegaOC}
Q_\Omega(s,o) = \sum_a \pi_{\theta_\pi}(a|s,o)Q_U(s,o,a),
\end{equation}
\vspace{-2mm}

$Q_U: \mathcal{S} \times \Omega \times \mathcal{A} \rightarrow \mathbb{R}$ is defined as the value of an action given the context of a state-option pair:

\vspace{-4mm}
\begin{equation} \label{QU}
Q_U(s,o,a) = r(s,a) + \gamma \sum_{s'} P(s'|s,a)U(s',o).
\end{equation}
\vspace{-3mm}

These $(s,o)$ pairs define an augmented state space \citep{Levy2011}. Instead, the option-critic architecture leverages the function $U: \Omega \times \mathcal{S} \rightarrow \mathbb{R}$ which is called the option-value function upon arrival \citep{Options}. The value of selecting option $o$ upon entering $s'$ is:

\vspace{-4mm}
\begin{equation} \label{U}
U(s',o) = ( 1 - \beta_{\theta_\beta}(s',o) ) Q_\Omega(s',o) + \beta_{\theta_\beta}(s',o) V_\Omega(s').
\end{equation}
\vspace{-3mm}

We adopt a notation for clarity where we omit $\theta_\pi$ and $\theta_\beta$ which $Q_U$ and $U$ both depend on. The \textit{intra-option policy gradient theorem} results from taking the derivative of the expected discounted return with respect to the intra-option policy parameters $\theta_\pi$, defining the update rule for $\pi$:

\vspace{-3mm}
\begin{equation}
\label{iopgt}
\sum_{s,o} \mu_\Omega(s,o|s_0,o_0) \sum_a \frac{\partial \pi_{\theta_\pi}(a|s,o)}{\partial \theta_\pi} Q_U(s,o,a).
\end{equation}
\vspace{-3mm}

$\mu_\Omega$ is defined as the discounted weighting of $(s,o)$ along trajectories originating from
$(s_0,o_0): \mu_\Omega(s,o|s_0,o_0) = \sum_{t=0}^{\infty} \gamma^t P(s_t=s,o_t=o|s_0,o_0)$. Likewise, the \textit{termination gradient theorem} results from taking the derivative of the expected discounted return with respect to the termination policy parameters $\theta_\beta$ and defines the update rule for $\beta$ with initial condition $(s_1,o_0)$:

\vspace{-4mm}
\begin{equation}
- \sum_{s',o} \mu_\Omega(s',o|s_1,o_o) \frac{\partial \beta_{\theta_\beta}(s',o)}{\partial \theta_\beta} A_\Omega(s',o),
\end{equation}
\vspace{-2mm}

where $A_\Omega$ is the advantage function over options, $A_\Omega(s',o) = Q_\Omega(s',o) - V_\Omega(s')$. 

\textbf{The hierarchical option-critic architecture} \citep{HOC} extends option-critic models 
to an arbitrarily deep $N$ level hierarchy of high level options and low level options below them. We adopt the notation from \citep{HOC} $x^{i:i+j} = x^i, ..., x^{i+j}$. This implies that  $x^{i:i+j}$ denotes a list of variables in the range of $i$ through $i+j$. In this hierarchical architecture,
$\pi_{\theta^1}^1(o^1|s)$ is the policy over the most abstract options in the hierarchy $o^1$. 
Once $o^1$ is chosen, 
$o^2$ is chosen with policy $\pi_{\theta^2}^2(o^2|s,o^1)$, which is the next highest level policy. 
This process continues 
until reaching policy $\pi_{\theta^N}^N(a|s,o^{1:N-1})$. $\pi^N$ is the lowest level policy and finally selects over the primitive action space. 
Each level of the option hierarchy also has a complimentary termination function $\beta_{\phi^1}^1(s,o^1),...,\beta_{\phi^{N-1}}^{N-1}(s,o^{1:N-1})$. 
Termination is bottom up, so high level options can only 
terminate when all lower level options have terminated first. 

At each level of abstraction $\ell$, the hierarchical option-critic architecture has an analogous option-value function $Q_\Omega(s,o^{1:\ell})$, value of selecting an option in the presence of previously selected options $Q_U(s,o^{1:\ell})$, and option-value function upon arrival $U(s,o^{1:\ell})$. The \textit{hierarchical intra-option policy gradient theorem} results from taking the derivative of the expected discounted return with respect to the policy parameters $\theta^\ell$, defining the update rule for $\pi^\ell$:

\vspace{-4mm}
\begin{equation}
\begin{split}
\sum_{s,o^{1:\ell}} \mu_\Omega(s,o^{1:\ell}|s_0,o_0^{1:N-1}) \frac{\partial \pi^\ell_{\theta^\ell}(o^\ell|s,o^{1:\ell-1})}{\partial \theta^\ell} Q_U(s,o^{1:\ell}),
\end{split}
\end{equation}
\vspace{-4mm}

where $\mu_\Omega$ is defined as the discounted weighting of $(s,o^{1:\ell})$ along trajectories originating from
$(s_0,o_0^{1:\ell}): \mu_\Omega(s,o^{1:\ell}|s_0,o_0^{1:N-1}) = \sum_{t=0}^{\infty} \gamma^t P(s_t=s,o_t^{1:\ell}=o^{1:\ell}|s_0,o_0^{1:N-1})$. The \textit{hierarchical termination gradient theorem} results from taking the derivative of the expected discounted return with respect to the termination policy parameters $\phi^\ell$ and defines the update rule for $\beta^\ell$ for the initial condition $(s_1,o_0^{1:N-1})$:

\vspace{-4mm}
\begin{equation}
\begin{split}
-\sum_{s,o^{1:\ell}}  \mu_\Omega(s,o^{1:\ell}|s_1,o_0^{1:N-1}) \prod_{i=\ell+1}^{N-1} \beta_{\phi^i}^i(s,o^{1:i}) \bigg( \\ \frac{\partial \beta_{\phi^\ell}^\ell(s,o^{1:\ell})}{\partial \phi^\ell} A_\Omega(s,o^{1:\ell}) \bigg),
\end{split}
\end{equation}
where $A_\Omega$ is the advantage function over a hierarchy options. 

\section{Policy Gradient Theorems Over A Full Option-Critic Architecture}

We now turn our attention to deriving policy gradient theorems for option-critic and hierarchical option-critic models by taking the derivative of the expected return with respect to a global set of shared parameters $\bm{\theta}$ rather than individually for each component at each level of abstraction.  

\subsection{Option-Critic Policy Gradients}

For the standard option-critic model, we follow the original paper and take the derivative of $Q_\Omega(s,o)$, but now with respect to $\bm{\theta}$ rather than the parameters of the different system components. This can be done by substituting in equation \ref{QOmegaOC}. 

\textbf{Lemma 1} (Option-Critic Policy Gradients). \textit{Given a set  of  Markov  options  with  stochastic policies and termination functions differentiable in their parameters $\bm{\theta}$ governing each option policy $\pi$, termination function $\beta$, and the policy over options $\pi_\Omega$, the gradient of the expected discounted return with respect to $\bm{\theta}$ and initial conditions $(s_0,o_0)$ is:}
\begin{footnotesize}
\begin{align*}
&\sum_{s,o,s'}\!\!\mu_\Omega(s,o,s'|s_0,o_0)\!\bigg(\!\sum_a \frac{\partial \pi(a|s,o)}{\partial \bm{\theta}} Q_U(s,o,a)  \\ &+\sum_{o'}\!\gamma \beta(s',o) \frac{\partial \pi_\Omega(o'|s')}{\partial \bm{\theta}} Q_\Omega(s',o')  - \gamma\frac{\partial \beta(s',o)}{\partial \bm{\theta}}A_\Omega(s',o)\bigg),
\end{align*}
\end{footnotesize}
where $\mu_\Omega$ is a discounted weighting of augmented state tuples along trajectories starting from $(s_0,o_0): \mu_\Omega(s,o,s'|s_0,o_0) = \sum_{t=0}^\infty \gamma^t P(s_t = s, o_t = o, s_{t+1} = s'|s_0,o_0)$. We provide one proof in Appendix A in the style of \citep{OC} and another complementary proof following \citep{kostas2019reinforcement} in Appendix C. An important point we are making here is that all option-critic style policy gradient theorems are just special cases of the more general co-agent policy gradient theorem \citep{kostas2019reinforcement}. In Appendix D we also provide a formal A3C style algorithm and computational analysis. Despite a more complex update, we theoretically and empirically show very similar computation time to prior option-critic models. 

Previously, \citet{baconphd} considered an almost identical policy gradient termed the \textit{joint gradient}. \citet{baconphd} did not include the discount factor for the next state, and no similar algorithms have actually been implemented in prior work to the best of our knowledge. However, we consider this result a lemma towards developing a more general theorem. It is not on its own a major theoretical result because it already existed and was first established by \citet{baconphd} and in a more general form by \citet{kostas2019reinforcement}. Our proof is not exactly the same as \citep{baconphd} mostly because it starts with a different equation for the value function. 

It is interesting that when using $\bm{\theta}$ we see the emergence of a single update rule with three separate terms that closely parallel the individual option-critic update rules for $\pi$, $\pi_\Omega$, and $\beta$. However, there are some considerable differences as well. There is now a formal acknowledgement of the policy over options and termination function being updated at the next state rather than the current state, which as such includes a multiplicative factor of $\gamma$. During training, in most cases this likely has a minor effect, especially if $\gamma$ is close to 1. 

\begin{figure}[t!]
    \centering
    \includegraphics[width=0.25\textwidth]{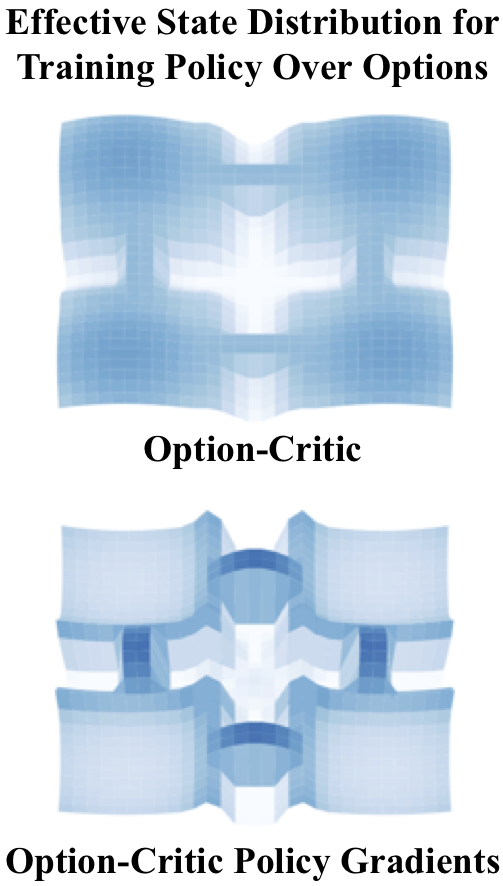}
    \caption{Illustration of effective state distribution for training $\pi_\Omega$ 
    with OCPG in four rooms environment. \emph{Top}: Typical option-critic learning focuses uniformly over states visited by the policy. \emph{Bottom}: OCPG transforms the distribution to focus on the transition points where options are likely to terminate and the policy over options is actually used.}
    \label{figure:state_distribution}
\end{figure}

The most influential new term is the multiplication by $\beta$ that is present in the update rule for the policy over options. This multiplicative factor is quite intuitive as it modulates the importance of updates to the policy over options by the likelihood that the current option terminates and the policy over options is actually used. Here, $\beta$ appears as a coupling factor between the dynamics of the policy over options and the dynamics of the underlying options being selected over. When the termination function is close to firing, the contribution of the update to the policy over options is the largest possible. In contrast, the smallest updates are obtained when the termination likelihood is close to zero. This term of the update therefore prioritizes likely transition points between options rather than treating all states equally. This is an interesting result as it shows that to strictly follow the exact policy gradient for the option-critic architecture, the policy over options cannot be viewed in total isolation from the underlying option dynamics as done in previous work by using simple Q-Learning or actor-critic. See Figure \ref{figure:state_distribution} for an illustration of how this refined policy gradient theorem promotes improved sample efficiency for learning the policy over options. 

The last difference that it is important to mention is the inclusion of $o'$ and not just $o$ in the update rule. One could be concerned about adding to the transition tuple needed for updates. This is not an issue at all for straightforward application of the option-critic policy gradient to on-policy learning. However, it may not prove useful for off-policy updates where the current policy is significantly different from the policy used in the environment. In those cases, this term can be removed from transition tuple stored and recovered by sampling from the current policy over options. Additionally, even for the $o$ term, it would probably be better to consider how the changing characteristics and current policy align with past behaviors as demonstrated by \citet{HiRO}. 

\subsection{A Generalized Hierarchical Option-Critic Policy Gradient Theorem}

In this section, following \citep{HOC}, we seek to find a generalization of the option-critic policy gradient to an arbitrarily deep option hierarchy with $N$ levels of abstraction. As we did in the last section, we start by taking the derivative of the option-value function for all active options $Q_\Omega(s,o^{1:N-1})$ with respect to $\bm{\theta}$. The derivation proceeds similarly to the one for the option-critic policy gradient. However, we now consider the complexities of an arbitrarily deep augmented state space of options. 

\textbf{Theorem 1} (Hierarchical Option-Critic Policy Gradient Theorem). \textit{Given an $N$ level hierarchical set  of  Markov  options  with  stochastic  option  policies at each level $\pi^\ell$ and termination functions at each level $\beta^\ell$ differentiable in their parameters $\bm{\theta}$, the gradient of the expected discounted return with respect to $\bm{\theta}$ and initial conditions $(s_0,o_0^{1:N-1})$ is: 
}
\begin{footnotesize}
\begin{align*}
    &\sum_{s,o^{1:N-1},s'}\!\!\!\!\mu_\Omega(s,o^{1:N-1},s'|s_0,o_0^{1:N-1})  \bigg( \\ &\sum_{a} \frac{\partial \pi(a|s,o^{1:N-1})}{\partial \bm{\theta}} Q_U(s,o^{1:N-1},a) + \gamma\!\!\!\!\sum_{o'^{1:N-1}} \sum_{\ell=1}^{N-1} \prod_{k=N-1}^{\ell} \bigg[ \\ &\!\!\!\!\beta^k(s',o^{1:k}) \frac{\partial \pi^\ell(o'^\ell|s',o'^{1:\ell-1})}{\partial \bm{\theta}} Q_\Omega(s',o'^{1:\ell}) P_{\pi,\beta}(o'^{1:\ell-1}|s',o^{1:\ell-1}) \bigg] \\ &- \gamma \sum_{\ell=1}^{N-1} \frac{\partial \beta^\ell(s',o^{1:\ell})}{\partial \bm{\theta}} A_\Omega(s',o^{1:\ell})\!\!\!\!\prod_{k=N-1}^{\ell+1}\!\!\!\!\beta^k(s',o^{1:k}) \bigg),
\end{align*}
\end{footnotesize}
where $\mu_\Omega$ is a discounted weighting of augmented states tuples 
along trajectories starting from $(s_0,o_0^{1:N-1}): \mu_\Omega(s,o^{1:N-1},s'|s_0,o_0^{1:N-1}) = \sum_{t=0}^\infty \gamma^t P(s_t=s, o_t^{1:N-1}=o^{1:N-1},s_{t+1}=s'|s_0,o_0^{1:N-1})$. $P_{\pi,\beta}(o'^{1:\ell-1}|s',o^{1:\ell-1})$ is the probability while at the next state and terminating the options for the current state that the agent arrives at a particular set of next option selections. We provide a proof in Appendix B in the style of \citep{HOC} and another proof in the style of \citep{kostas2019reinforcement} in Appendix C. We also provide a formal A3C style algorithm and analysis theoretically showing very similar computation per step to prior hierarchical option-critic models in Appendix D. 

We first would like to point out that this theorem is a generalization of the option-critic policy gradient in the previous section. In fact, when $N=2$ the hierarchical option-critic policy gradient theorem should be exactly the same. \footnote{Note that $P_{\pi,\beta}=1$ when $N=2$.} In comparison to the original hierarchical intra-option policy gradient theorem and hierarchical termination gradient theorem from \citep{HOC}, our new theorem has many of the same terms but with only one update rule rather than $2N-1$ different update rules. The other chief differences with the original work are similar to the last section in that option policies and termination functions update with respect to the next state. Additionally, we again see the emergence of a dependence of the option policy update rules on the likelihood that the option policy is actually used. The option policy is only used if the both the option at that level of abstraction and all lower level options terminate.




\section{Empirical Analysis}

We now seek to evaluate how OCPG and HOCPG perform in a function approximation setting with a complex state space and thus consider the Atari games \citep{ALE}. We utilize the popular Open AI Gym environments for these games and use the default settings. We extend A3C from a popular PyTorch repository and provide further details on our setup in Appendix E. Our architecture follows \citet{DQN} consisting of a feature extractor common across all components of the architecture with 4 convolutional layers each followed by a max pooling and ReLU layer. This output is then fed into an LSTM as in \citep{A3C}. 

\textbf{Implementation Details:} We implement option-critic policy gradients (OCPG) using the variant of A2OC outlined in algorithm 1 of Appendix D and implement hierarchical option-critic policy gradients (HOCPG) following algorithm 2 of Appendix D. Our primary baselines (OC) and (HOC) are standard version of A2OC and A2HOC respectively using the intra-option policy gradient theorem and termination gradient theorem for training. For easier direct comparison with our new full architecture level policy gradient theorem, we leverage actor-critic learning for the policy over options rather than arbitrarily using Q-Learning. This allows us to directly compare the ability of models to implement a policy gradient theorem for the full system.  All of our models use 8 options following past work, and a learning rate of 1e-4.  Following \citep{Deliberation} we run 16 parallel asynchronously updating threads for each game. We also run one thread for evaluation that we do not learn from. We report the average and standard deviation of the reward for the most recent 150 evaluation episodes across ten runs. To ensure that our analysis is statistically sound, given the high variance that is typical for deep reinforcement learning, we follow best practices from \citep{stats}. 

We should also note that it was the aim of prior work to study how and when options are useful. Our paper instead focuses on a particular systemic and problem agnostic aspect of the optimization of option-critic learning. As such, in the main text we only have space to report our main results. However, we have provided additional analysis of the options that are actually learned by each model and where potential learning advantages may be coming from in Appendix E. 

\begin{figure*}[t!]
  \centering
    \includegraphics[width=0.75\textwidth]{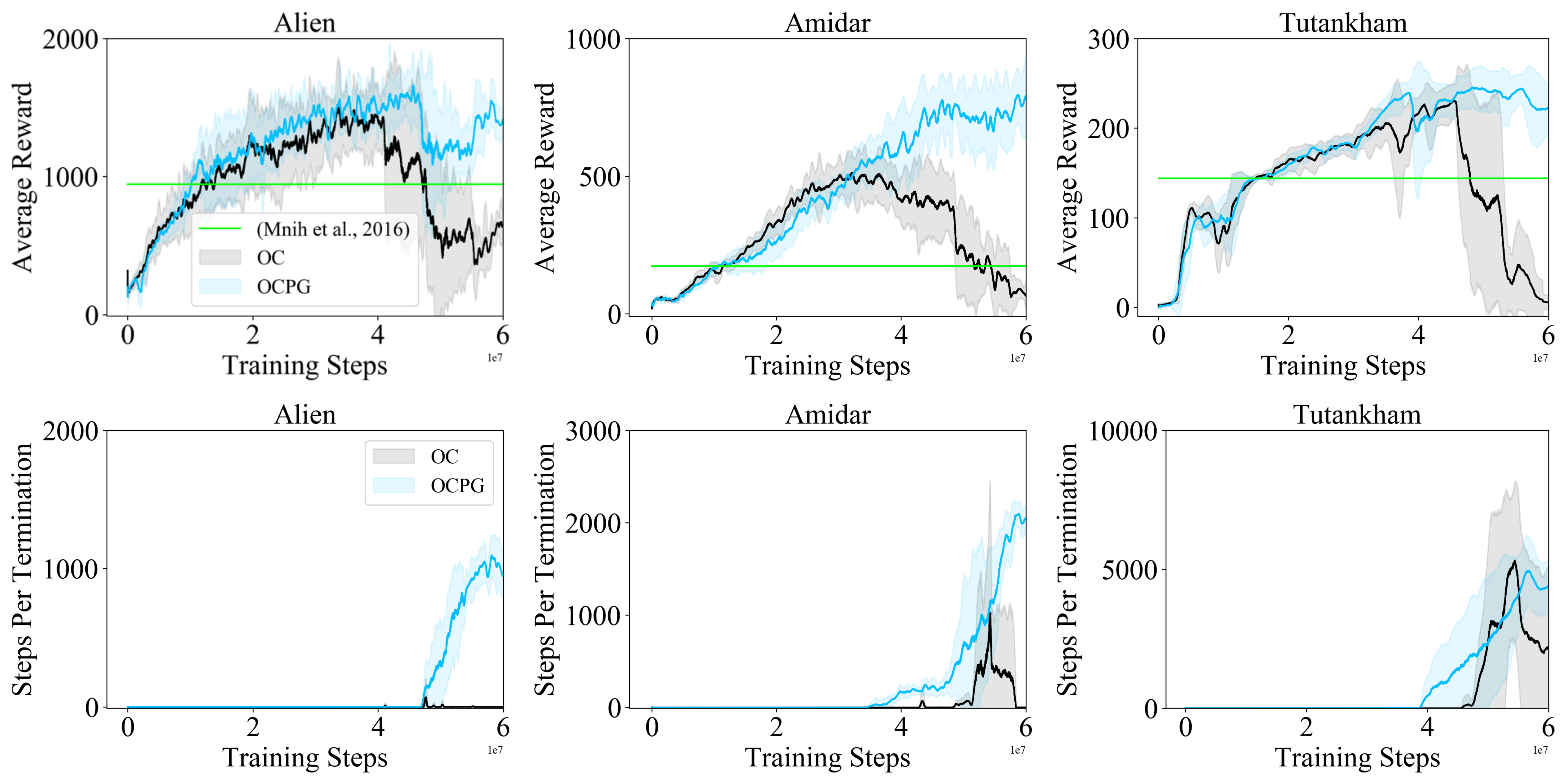}
    \caption{Option-critic style model performance and steps per termination during learning with a $\eta$ schedule.} \label{OCAtari}
\end{figure*}

\textbf{Settings Explored:} It is important to note that Lemma 1 and Theorem 1 do not differ significantly from past work on option-critic and hierarchical option-critic learning for the corner case where all options always terminate. As such, it is clear that the gap with old approaches for option-critic learning becomes more substantial as options become more extended in time. On the other hand, this fact poses a challenge for conducting experiments that highlight the value gained as a result of the modified gradient. This is because option-critic style architectures since the first paper \citep{OC} have needed to regularize the advantage function during the updates to the termination function using a parameter $\eta$ to avoid trivially learning to always terminate. Option-critic style models paired with a regularizer can have the ability to learn options ranging the spectrum of temporal lengths, but have also been shown to be quite sensitive to the value of this parameter \citep{Deliberation}. In our experiments, we attempt to understand the role that regularization and option termination frequency have on the behavior of OCPG and OC. To do this we consider two settings of interest. In the first setting we consider a simple schedule (similar to a learning rate schedule) for $\eta$ in which OCPG and HOCPG perform quite well. Then we explore choosing a reasonable fixed $\eta$ that leads to options that are both extended in time and still divide the episode into segments.  

\subsection{Regularization Schedule}

\begin{figure}[!b]
  \centering
    \includegraphics[width=0.5\textwidth]{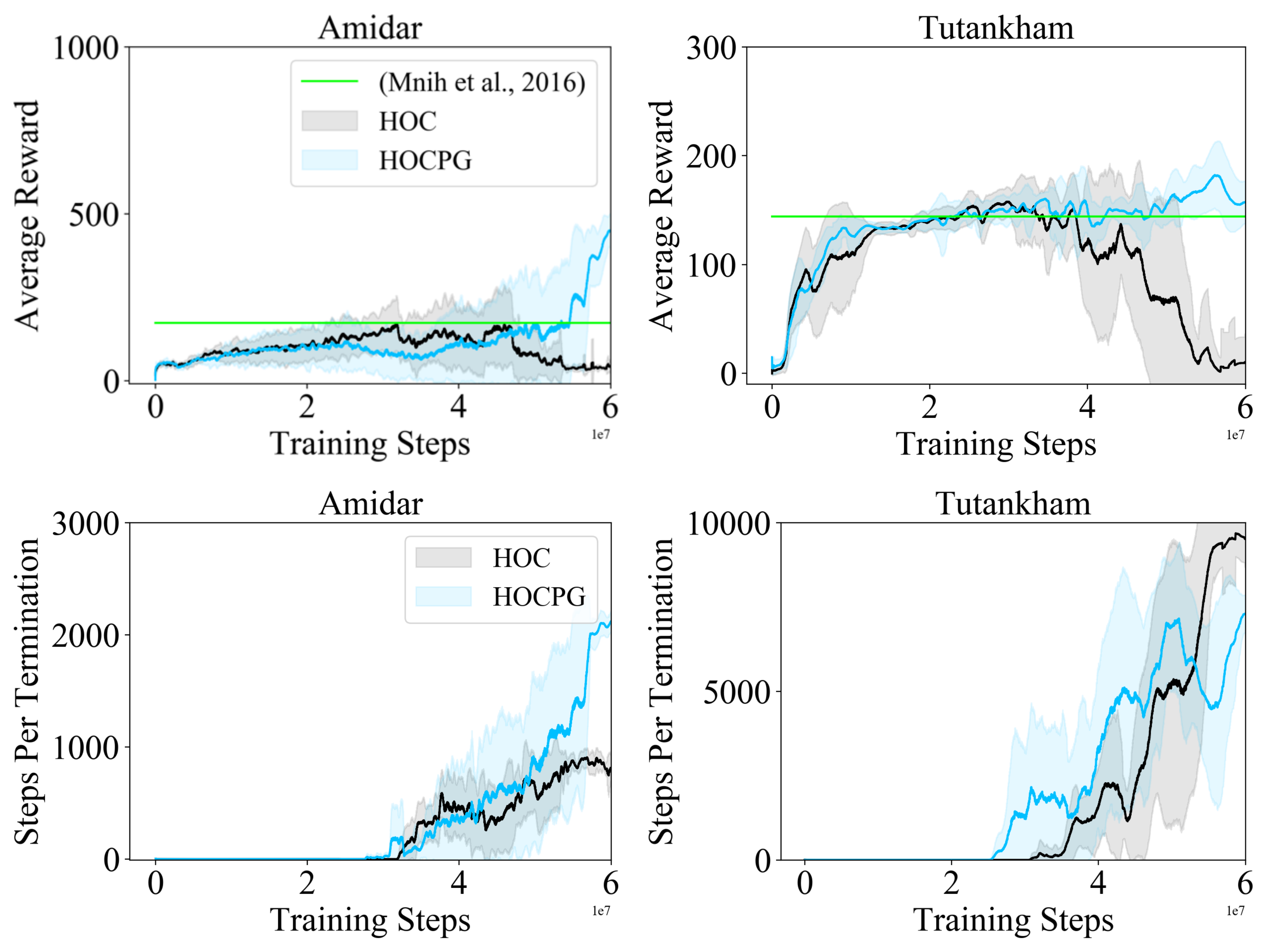}
    \caption{Hierarchical option-critic style model performance and steps per termination of the highest level option during learning with a $\eta$ schedule.}  \label{HOCAtari}
\end{figure}

\begin{figure*}[t!]
  \centering
    \includegraphics[width=0.75\textwidth]{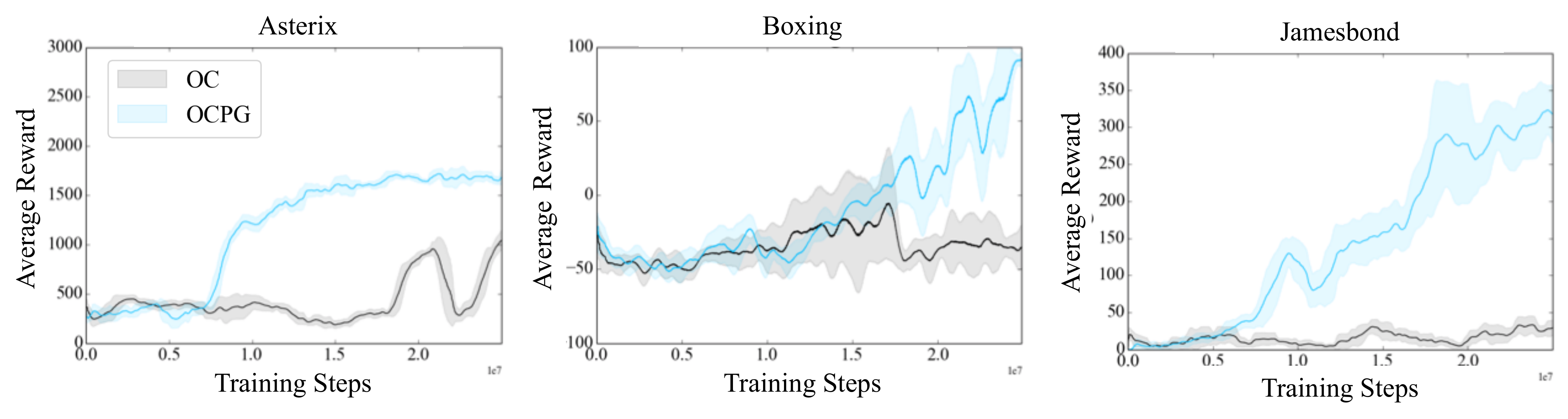}
    \caption{Option-critic style model performance for three Atari games with fixed $\eta=0.3$. } \label{LearningCurves}
\end{figure*}

While high values of $\eta$ lead to options that are of longer duration, they have the drawback of leading to solutions that are merely $\eta$-optimal \citep{Deliberation}. This implies that we would only arrive at an optimal policy for the base MDP if $\eta=0$. With this in mind, it seems quite natural to set a schedule for an agent trying to solve a task where first the agent focuses on learning skills from scratch without temporal abstraction and then the agent gradually learns to terminate less, decomposing the problem in time. Before any useful policies are learned, the regularization on the advantage may only serve to impede the proper convergence of the model. To test out this theory, we consider Atari experiments with training for 60 million steps while implementing a simple regularization schedule where $\eta$ changes every 15 million steps. At the beginning we set $\eta=0.0$ for 15 million steps before setting $\eta=0.01$. Next, we set $\eta=0.1$ at 30 million steps and $\eta=1.0$ at 45 million steps. We tested six runs of each algorithm in this setup on three maze style navigation environments: Alien, Amidar, and Tutankham. For the hierarchical models, we set $\eta$ to the same value at each level.    


In Figure \ref{OCAtari}, we report results for OCPG and OC across Alien, Amidar, and Tutankham for 5 random runs. We include A3C performance from Table S3 of \citep{A3C} as a point of reference for how agents with primitive actions perform on these games. We report the results of a similar agent with a CNN and LSTM based representation like ours, but we should note this results are not directly comparable in a few ways. For example, the agents in past work were trained for considerably more steps. 
In early training, OCPG and OC have pretty much identical performance. This is particularly noticeable when $\eta=0$ during the first 15 million steps. However, as $\eta$ gradually increases, we begin to see OCPG significantly differentiate itself from OC. OC experiences significant instability once $\eta=1.0$ at 45 million steps of training in all three games. Meanwhile, OCPG does not really see this instability or at least experiences it much less. Below our average reward results, we also provide insight about the termination behavior of the agents by plotting the average number of steps per termination over time. We can see that the gap between OCPG and OC consistently begins for each game when we start to see terminations become less frequent. Inline with our remarks in the last section, it is predictable that both algorithms should have the same performance when all options terminate every step. Additionally, it makes perfect sense that the gap between algorithms grows as $\eta$ becomes high and options start taking a significant amount of time. When options are longer, OCPG has a superior ability to focus the policy over options on likely transition regions, allowing it to quickly adapt to and manage longer options.   

We also report results for hierarchical option models in Figure \ref{HOCAtari} when using $N=3$ levels of abstraction (standard option-critic uses $N=2$). Unfortunately, both hierarchical models struggled to achieve the performance of a primitive action agent on Alien (not shown) and didn't seem to provide additional value over option-critic on the other games by using the third level of abstraction either (at least with only 60 million steps of training from scratch). Nonetheless, the model still surpasses prior results with primitive actions in a short amount of training and can still be useful for understanding the optimization behavior of the hierarchical option-critic architecture. Similar to the $N=2$ case, we see significant instability in the HOC model that we do not see for HOCPG. Once again, in line with what we would expect, we see that the gap between the models only starts to grow when options start learning not to terminate. A noticeable difference with our option-critic experiments is that options become longer earlier. However, this could be related to our setting as $\eta$ grows in effect when applied to more levels of abstraction, making the effective regularization higher at the same $\eta$. 

\subsection{Fixed Regularization}

We would also like to verify that our proposed algorithms add value in the typical setting where $\eta$ is simply set to some reasonable value. Based on our initial experiments we found that $\eta=0.3$ was a good choice that led to options of non-trivial length for each game we explored. In addition to our three games from the previous section, we will explore four other Atari games that have quite different dynamics than these maze style navigation games including Asterix, Boxing, Jamesbond, and Tennis. We consider a sample efficient setting similar to \citep{NEC} where we evaluate model performance after 10 million and 25 million training steps. This allows us to highlight the improved learning efficiency of OCPG when options are of non-trivial lengths. 

In Table \ref{Atari7Games} we report results for OCPG and OC across 10 runs. While, again, results may not be directly comparable due to differences in the implementation, past work has reported performance for primitive action A3C after 10 million steps as 415.5 for Alien, 96.3 for Amidar, 301.6 for Asterix, 2.5 for Boxing, 31.5 for Jamesbond, -23.8 for Tennis, and 108.3 for Tutankham \citep{NEC}. In all but one case, OCPG results in superior performance to both OC and prior reported A3C results. The only apparent failure is Boxing at 10 million steps. That said, the performance achieved for Boxing at 25 million steps is very impressive given the amount of training and even surpasses some past asymptotic results reported for A3C \citep{A3C}. Once again, we find that the strong comparative performance of OCPG is in a setting where options often do not terminate. For example, even by 10 million training steps, the average number of terminations per step for OCPG is 7.9 for Alien, 12.8 for Amidar, 8.3 for Asterix, 32.4 for Boxing, 5.5 for Jamesbond, 242.2 for Tennis, and 1266.3 for Tutankham. 

\begin{table}
\centering
\setlength{\tabcolsep}{2pt}
\resizebox{1.0\columnwidth}{!}{
\begin{tabular}{|c|c|c|}
\hline
Game & OC & OCPG  \\
& \small 10M / 25M & \small 10M / 25M \\ \hline
Alien & 491.5 \small{$\pm$ 115.3} \normalsize / 618.8 \small{$\pm$ 61.8} & \textbf{663.0} \small{$\pm$ 45.3} \normalsize \normalsize / \textbf{791.4} \small{$\pm$ 44.7} \\ \hline
Amidar & 45.1 \small{$\pm$ 26.4} \normalsize / 89.4 \small{$\pm$ 15.9} & \textbf{219.3} \small{$\pm$ 18.4} \normalsize / \textbf{141.9} \small{$\pm$ 18.6} \\ \hline
Asterix  & 409.7 \small{$\pm$ 76.6} \normalsize / 1042.0 \small{$\pm$ 98.1} & \textbf{1213.4} \small{$\pm$ 74.3} \normalsize / \textbf{1679.8} \small{$\pm$ 71.0} \\ \hline
Boxing  & -37.7 \small{$\pm$ 14.7} \normalsize / 32.9 \small{$\pm$ 14.3} & -41.4 \small{$\pm$ 12.9} \normalsize / \textbf{90.8} \small{$\pm$ 6.8} \\ \hline
Jamesbond & 7.1 \small{$\pm$ 5.5} \normalsize / 29.3 \small{$\pm$ 10.7} & \textbf{118.6} \small{$\pm$ 29.4} \normalsize / \textbf{317.7} \small{$\pm$ 38.7} \\ \hline 
Tennis & -17.9 \small{$\pm$ 2.6} \normalsize / -19.9 \small{$\pm$ 1.3} & \textbf{-14.4} \small{$\pm$ 1.7} \normalsize / \textbf{-13.1} \small{$\pm$ 2.8} \\ \hline  
Tutankham & 1.6 \small{$\pm$ 2.5} \normalsize / 0.8 \small{$\pm$ 3.0} & \textbf{16.8} \small{$\pm$ 20.8} \normalsize / \textbf{128.0} \small{$\pm$ 27.3} \\ \hline
\end{tabular}}
\caption{ Average and standard deviation of the evaluation reward for Atari games across 10 runs, reported after 10 million steps (10M) and 25 million steps (25M). Bold indicates a statistically significant difference according to Welch's t-test.}
\label{Atari7Games}
\end{table}

In Figure \ref{LearningCurves}, we highlight the learning behavior of OCPG and OC on three of the new games. For Asterix and Jamesbond, we see OCPG pretty immediately differentiate itself from the OC baseline and achieve quite impressive early performance. However, OCPG takes a little longer to differentiate itself from OC on Boxing. That being said, once it starts learning, it is able to achieve great performance quickly. 

\section{Discussion}

Developing better general purpose methods for learning options from scratch is an important avenue of RL research. In this work, we reconsider an assumption in the intra-option policy gradient theorem and termination gradient theorem, namely that the policies and termination functions have independent parameters \citep{OC}. While this assumption is true in tabular settings, it is never the case for practical deep function approximation settings as parameter sharing across components of the model leads to sample efficient learning. We propose to rectify this issue by performing full architecture level \emph{option-critic policy gradient} updates. A key distinction with prior approaches is that the update rule for the policy over options depends on the behavioral characteristics of the underlying options it selects over. We demonstrate that this modification leads to improved sample efficient learning across a test bed of Atari games. We show for a number of games that this update rule results in more stable and faster learning by focusing on a more representative state space distribution when options are long. 



\section*{Acknowledgements}
The authors thank Murray Campbell and Tim Klinger for fruitful discussions that helped shape this work. We also would like to thank Modjtaba Shokrian-Zini and Akshay Dharmavaram for their helpful comments and suggestions. Additionally, we thank Dan Jurafsky, Chris Potts, and Josh Greene for their support.

\bibliographystyle{iclr2019}
\bibliography{example_paper}

\appendix

\section{Appendix A: Proof of Option-Critic Policy Gradient Theorem} \label{OCPGProof}

\subsection{Remarks About the Augmented Process}

Directly following the original option-critic policy gradient theorem proof \citep{OC}, we begin by making a remark about augmented processes which will be helpful later in the final steps of the derivation. If $o_t$ is executing at time $t$, then the discounted probability of transitioning to $(s_{t+1},o_{t+1})$ is:

\begin{equation} \label{P1}
\begin{split}
P_\gamma^{(1)}(s_{t+1},o_{t+1}|s_t,o_t) = \sum_a \pi(a|s_t,o_t) \gamma P(s_{t+1}|s_t,a) \bigg( \\ (1 - \beta(s_{t+1},o_t))\textbf{1}_{o_t=o_{t+1}} + \beta(s_{t+1},o_t) \pi_\Omega(o_{t+1}|s_{t+1}) \bigg).
\end{split}
\end{equation}

The discounted probabilities for k-steps can more generally be expressed recursively:

\begin{align}
\begin{split}
P_\gamma^{(k)}(s_{t+k},o_{t+k}|s_t,o_{t}) =
\sum_{s_{t+1},o_{t+1}} \bigg( \\ P_\gamma^{(1)}(s_{t+1},o_{t+1}|s_t,o_{t}) P_\gamma^{(k-1)}(s_{t+k},o_{t+k}|s_{t+1},o_{t+1}) \bigg).
\end{split}
\end{align}

\subsection{Derivation of Lemma 1}

We begin by following the proof of the intra-option policy gradient theorem from \citet{OC}. We take the gradient of the option-value function now with respect to $\bm{\theta}$ rather than just $\theta_\pi$:

\begin{equation}
\begin{split}
\frac{\partial Q_\Omega(s,o)}{\partial \bm{\theta}} = \frac{\partial}{\partial \bm{\theta}} \sum_a \pi(a|s,o) Q_U(s,o,a) \\ 
= \sum_a \left(\frac{\partial \pi(a|s,o)}{\partial \bm{\theta}} Q_U(s,o,a) + \pi(a|s,o) \frac{\partial Q_U(s,o,a)}{\partial \bm{\theta}}\right).
\end{split}
\end{equation}

Substituting in equation 2 of the main text
, we continue to expand our expression:

\begin{equation} \label{QPart}
\begin{split}
\frac{\partial Q_\Omega(s,o)}{\partial \bm{\theta}} = \sum_a \bigg(\frac{\partial \pi(a|s,o)}{\partial \bm{\theta}} Q_U(s,o,a) \\ + \pi(a|s,o) \sum_{s'} \gamma P(s'|s,a) \frac{\partial U(s',o)}{\partial \bm{\theta}}\bigg).
\end{split}
\end{equation}

Still following the proof from \citet{OC}, we seek to further expand the last term in this expression.  
However, this now leads to new terms that were not considered in the original work as we are taking the derivative with respect to $\bm{\theta}$ and not only $\theta_\pi$. We then take the derivative of equation 3 of the main text 
to understand the influence of that term on equation \ref{QPart}:

\begin{equation} 
\begin{split}
\frac{\partial U(s',o)}{\partial \bm{\theta}} &= (1-\beta(s',o)) \frac{\partial Q_\Omega(s',o)}{\partial \bm{\theta}} + \beta(s',o)\frac{\partial V_\Omega(s')}{\partial \bm{\theta}} \\ &+ \frac{\partial \beta(s',o)}{\partial \bm{\theta}}V_\Omega(s') -\frac{\partial \beta(s',o)}{\partial \bm{\theta}}Q_\Omega(s',o).
\end{split}
\end{equation}

We can then substitute in the advantage function over options $A_\Omega(s',o) = Q_\Omega(s',o) - V_\Omega(s')$:

\begin{align} \label{eqU} 
\begin{split}
\frac{\partial U(s',o)}{\partial \bm{\theta}} &= (1-\beta(s',o)) \frac{\partial Q_\Omega(s',o)}{\partial \bm{\theta}} \\ &+ \beta(s',o)\frac{\partial V_\Omega(s')}{\partial \bm{\theta}} -\frac{\partial \beta(s',o)}{\partial \bm{\theta}}A_\Omega(s',o).
\end{split}
\end{align}

We can define the value function $V_\Omega(s')$ by in terms of the option-value function using the policy over options:

\begin{equation} 
\begin{split}
V_\Omega(s') = \sum_{o'} \pi_\Omega(o'|s') Q_\Omega(s',o').
\end{split}
\end{equation}

Then we can take the gradient to obtain:
\begin{equation}
\begin{split} 
\frac{\partial V_\Omega(s')}{\partial \bm{\theta}} = \sum_{o'} &\bigg( \frac{\partial \pi_\Omega(o'|s')}{\partial \bm{\theta}} Q_\Omega(s',o') \\ &+ \pi_\Omega(o'|s') \frac{\partial Q_\Omega(s',o')}{\partial \bm{\theta}} \bigg).
\end{split}
\end{equation}

This allows us to expand our expression for equation \ref{eqU}: 

\begin{align} 
\begin{split}
\frac{\partial U(s',o)}{\partial \bm{\theta}} = (1-\beta(s',o)) \frac{\partial Q_\Omega(s',o)}{\partial \bm{\theta}} \\+ \beta(s',o)\sum_{o'} \bigg( \frac{\partial \pi_\Omega(o'|s')}{\partial \bm{\theta}} Q_\Omega(s',o') \\+ \pi_\Omega(o'|s') \frac{\partial Q_\Omega(s',o')}{\partial \bm{\theta}} \bigg) -\frac{\partial \beta(s',o)}{\partial \bm{\theta}}A_\Omega(s',o).
\end{split}
\end{align}

Substituting back into equation \ref{QPart} we now arrive at the following expression:

\begin{align}
\begin{split}
\frac{\partial Q_\Omega(s,o)}{\partial \bm{\theta}} = \sum_a \bigg(\frac{\partial \pi(a|s,o)}{\partial \bm{\theta}} Q_U(s,o,a) \\ + \pi(a|s,o) \sum_{s'} \gamma P(s'|s,a) \bigg[ \\
(1-\beta(s',o)) \frac{\partial Q_\Omega(s',o)}{\partial \bm{\theta}} + \beta(s',o)\sum_{o'} \big[ \frac{\partial \pi_\Omega(o'|s')}{\partial \bm{\theta}} Q_\Omega(s',o') 
\\ + \pi_\Omega(o'|s') \frac{\partial Q_\Omega(s',o')}{\partial \bm{\theta}} \big]
-\frac{\partial \beta(s',o)}{\partial \bm{\theta}}A_\Omega(s',o)\bigg]\bigg).
\end{split}
\end{align}

We can then reformulate this expression in order to make the recursive dynamics more clear:

\begin{equation}
\begin{split}
\frac{\partial Q_\Omega(s,o)}{\partial \bm{\theta}} = \sum_a \bigg(\frac{\partial \pi(a|s,o)}{\partial \bm{\theta}} Q_U(s,o,a) \\ + \pi(a|s,o) \sum_{s'} \sum_{o'} \gamma P(s'|s,a) \bigg[ (1-\beta(s',o)) \textbf{1}_{o'=o} \\ + \beta(s',o) \pi_\Omega(o'|s') \bigg] \frac{\partial Q_\Omega(s',o')}{\partial \bm{\theta}}  + \pi(a|s,o) \sum_{s'} \gamma P(s'|s,a) \bigg[ \\ \beta(s',o)\sum_{o'}  \frac{\partial \pi_\Omega(o'|s')}{\partial \bm{\theta}} Q_\Omega(s',o') -\frac{\partial \beta(s',o)}{\partial \bm{\theta}}A_\Omega(s',o) \bigg] \bigg).
\end{split}
\end{equation}

This equation can then be further simplified by substituting in equation \ref{P1}: 

\begin{equation}
\begin{split}
\frac{\partial Q_\Omega(s,o)}{\partial \bm{\theta}} = \sum_a \frac{\partial \pi(a|s,o)}{\partial \bm{\theta}} Q_U(s,o,a)  \\ +  \sum_{s'} \sum_{o'} P^{(1)}_\gamma(s',o'|s,o) \frac{\partial Q_\Omega(s',o')}{\partial \bm{\theta}} \\ + \sum_a \pi(a|s,o) \sum_{s'} \gamma P(s'|s,a) \bigg( \\ \sum_{o'} \beta(s',o) \frac{\partial \pi_\Omega(o'|s')}{\partial \bm{\theta}} Q_\Omega(s',o')  -\frac{\partial \beta(s',o)}{\partial \bm{\theta}}A_\Omega(s',o) \bigg).
\end{split}
\end{equation}

We can reorder and simplify this expression using the fact that $P(s'|s,o) = \sum_a\pi(a|s,o)P(s'|s,a)$:

\begin{equation}
\begin{split}
\frac{\partial Q_\Omega(s,o)}{\partial \bm{\theta}} = \sum_a \frac{\partial \pi(a|s,o)}{\partial \bm{\theta}} Q_U(s,o,a) \\ + \sum_{s'} \gamma P(s'|s,o) \bigg( \sum_{o'} \beta(s',o) \frac{\partial \pi_\Omega(o'|s')}{\partial \bm{\theta}} Q_\Omega(s',o')  \\ -\frac{\partial \beta(s',o)}{\partial \bm{\theta}}A_\Omega(s',o) \bigg). + \sum_{s'} \sum_{o'} P^{(1)}_\gamma(s',o'|s,o) \frac{\partial Q_\Omega(s',o')}{\partial \bm{\theta}}
\end{split}
\end{equation}

We can also rearrange this expression for simplicity by multiplying the first term by $\sum_{s'} P(s'|s,o)=1$ as the term does not depend on $s'$:

\begin{equation}
\begin{split}
\frac{\partial Q_\Omega(s,o)}{\partial \bm{\theta}} = \sum_a \sum_{s'} P(s'|s,o) \bigg( \frac{\partial \pi(a|s,o)}{\partial \bm{\theta}} Q_U(s,o,a) \\ + \gamma \sum_{o'} \beta(s',o) \frac{\partial \pi_\Omega(o'|s')}{\partial \bm{\theta}} Q_\Omega(s',o')  - \gamma \frac{\partial \beta(s',o)}{\partial \bm{\theta}}A_\Omega(s',o) \bigg). \\ +\sum_{s'} \sum_{o'} P^{(1)}_\gamma(s',o'|s,o) \frac{\partial Q_\Omega(s',o')}{\partial \bm{\theta}}
\end{split}
\end{equation}

Using the structure of the augmented process from the previous section and the fact that $P_\gamma^{(k)}(s',o',s''|s,o) = P(s''|s',o') P_\gamma^{(k)}(s',o'|s,o)$
we finally obtain:

\begin{equation}
\begin{split}
\frac{\partial Q_\Omega(s,o)}{\partial \bm{\theta}} = \sum_{s',o',s''} \sum_{k=0}^\infty P^{(k)}_\gamma(s',o',s''|s,o) \bigg( \\ \sum_{a'} \frac{\partial \pi(a'|s',o')}{\partial \bm{\theta}} Q_U(s',o',a') \\ + \sum_{o''} \gamma \beta(s'',o') \frac{\partial \pi_\Omega(o''|s'')}{\partial \bm{\theta}} Q_\Omega(s'',o'') \\ - \gamma \frac{\partial \beta(s'',o')}{\partial \bm{\theta}}A_\Omega(s'',o') \bigg).
\end{split}
\end{equation}

Therefore the gradient of the expected discounted return with respect to $\bm{\theta}$ is:

\begin{equation}
\begin{split}
\frac{\partial Q_\Omega(s_0,o_0)}{\partial \bm{\theta}} = \sum_{s,o,s'} \sum_{k=0}^\infty P^{(k)}_\gamma(s,o,s'|s_0,o_0)  \bigg( \\ \sum_{a} \frac{\partial \pi(a|s,o)}{\partial \bm{\theta}} Q_U(s,o,a) \\ + \sum_{o'} \gamma \beta(s',o) \frac{\partial \pi_\Omega(o'|s')}{\partial \bm{\theta}} Q_\Omega(s',o') \\ - \gamma \frac{\partial \beta(s',o)}{\partial \bm{\theta}}A_\Omega(s',o) \bigg) \\
= \sum_{s,o,s'} \mu_\Omega(s,o,s'|s_0,o_0) \bigg( \sum_{a}  \frac{\partial \pi(a|s,o)}{\partial \bm{\theta}} Q_U(s,o,a) \\ + \sum_{o'} \gamma \beta(s',o) \frac{\partial \pi_\Omega(o'|s')}{\partial \bm{\theta}} Q_\Omega(s',o') \\ - \gamma \frac{\partial \beta(s',o)}{\partial \bm{\theta}}A_\Omega(s',o) \bigg).
\end{split}
\end{equation}

\section{Appendix B: Proof of Hierarchical Option-Critic Policy Gradient Theorem} \label{HOCPGProof}

\subsection{Remarks About the Augmented Process}

Directly following the original hierarchical option-critic policy gradient theorem proof \citep{HOC}, we begin by making a remark about augmented processes which will be helpful later in the final steps of the derivation. If $o_t^{1:N-1}$ is executing at time $t$, then the discounted probability of transitioning to $(s_{t+1},o_{t+1}^{1:N-1})$ is:
\begin{equation} \label{P1HOC}
\begin{split}
P_\gamma^{(1)}(s_{t+1},o_{t+1}^{1:N-1}|s_t,o_t^{1:N-1}) = \\ \sum_a \pi(a|s_t,o_t^{1:N-1}) \gamma P(s_{t+1}|s_t,a) \bigg( \\ (1 - \beta^{N-1}(s_{t+1},o_t^{1:N-1}))\textbf{1}_{o_{t+1}^{1:N-1} =o_t^{1:N-1}} \\ + \prod_{j=N-1}^{1} \beta^j(s_{t+1},o_t^{1:j}) \pi^j(o_{t+1}^j|s_{t+1},o_{t+1}^{1:j-1}) \\ +
\sum_{i=1}^{N - 2}(1-\beta^i(s',o^{1:i})) \textbf{1}_{o_{t+1}^{1:i}=o_t^{1:i}} \bigg[ \\ \prod_{k=N-1}^{i+1} \beta^k(s',o_t^{1:k}) \pi^k(o_{t+1}^k|s_{t+1},o_{t+1}^{1:k-1}) \bigg] \bigg).
\end{split}
\end{equation}

The discounted probabilities for k-steps can more generally be expressed recursively: 

\begin{equation} 
\begin{split}
P_\gamma^{(k)}(s_{t+k},o_{t+k}^{1:N-1}|s_t,o_{t}^{1:N-1}) = \\ \sum_{s_{t+1},o_{t+1}^{1:N-1}} \bigg( P_\gamma^{(1)}(s_{t+1},o_{t+1}^{1:N-1}|s_t,o_{t}^{1:N-1}) \\ P_\gamma^{(k-1)}(s_{t+k},o_{t+k}^{1:N-1}|s_{t+1},o_{t+1}^{1:N-1}) \bigg).
\end{split}
\end{equation}

\subsection{Derivation of Theorem 1}

We now extend our analysis in the last section for the standard two-level option-critic architecture to an architecture with $N$ levels of abstraction as in \citet{HOC}. In order to get started, we must first define $Q_\Omega$, $Q_U$, and $U$ for state $s$ and active options $o^{1:N-1}$ directly following \citet{HOC}. The option-value function $Q_\Omega$ can be expressed as:

\begin{equation} 
Q_\Omega(s,o^{1:N-1}) = \sum_{a} \pi^N(a|s,o^{1:N-1})Q_U(s,o^{1:N-1},a).
\end{equation}

Likewise, the value of executing an action in the presence of the currently active options $Q_U$ can be expressed as: 

\begin{equation} 
\begin{split}
Q_U(s,o^{1:N-1},a) = r(s,a) + \gamma \sum_{s'} P (s'|s,a)U(s',o^{1:N-1}).
\end{split}
\end{equation}

We also follow the option value function upon arrival $U$ from \citet{HOC}. As we are focusing on the full active option hierarchy, we do not need the term for only lower level options terminating:   

\begin{equation} 
\begin{split}
U(s',o^{1:N-1}) = \underbrace{(1-\beta^{N-1}(s',o^{1:N-1}))Q_\Omega(s',o^{1:N-1})}_{\text{none terminate ($N \geq 1$)}} \\ + \underbrace{V_\Omega(s') \prod_{j=N-1}^{1} \beta^j(s',o^{1:j})}_{\text{all options terminate ($N \geq 2$)}} \\ +
\underbrace{ \sum_{i=1}^{N - 2}(1-\beta^i(s',o^{1:i})) Q_\Omega(s',o^{1:i}) \prod_{k=N-1}^{i+1} \beta^k(s',o^{1:k}) }_{\text{some options terminate ($N \geq 3$)}}.
\end{split}
\end{equation}

We can now follow a similar procedure to the one explored in the first section of the appendix, taking the derivative of $Q_\Omega(s,o^{1:N-1})$ with respect to $\bm{\theta}$: 

\begin{equation}  \label{QOmega}
\begin{split}
\frac{\partial Q_\Omega(s,o^{1:N-1})}{\partial \bm{\theta}} = \\ \frac{\partial}{\partial \bm{\theta}}\sum_{a} \pi^N(a|s,o^{1:N-1})Q_U(s,o^{1:N-1},a) \\ 
= \sum_{a} \frac{ \partial \pi^N(a|s,o^{1:N-1})}{\partial \bm{\theta}} Q_U(s,o^{1:N-1},a)
\\ + \sum_{a} \pi^N(a|s,o^{1:N-1}) \frac{\partial Q_U(s,o^{1:N-1},a)}{\partial \bm{\theta}},
\end{split}
\end{equation}

To further understand this equation, we also need to take the derivative of $Q_U(s,o^{1:N-1},a)$ with respect to $\bm{\theta}$:

\begin{equation} 
\begin{split}
\frac{ \partial Q_U(s,o^{1:N-1},a)}{\partial \bm{\theta}} = \gamma \sum_{s'} P (s'|s,a) \frac{\partial U(s',o^{1:N-1})}{\partial \bm{\theta}},
\end{split}
\end{equation}

To expand this expression, we need to take the derivative of $U(s',o^{1:N-1})$ with respect to $\bm{\theta}$ as well:

\begin{equation} \label{eq31}
\begin{split}
\frac{\partial U(s',o^{1:N-1})}{\partial \bm{\theta}} = -\frac{ \partial \beta^{N-1}(s',o^{1:N-1})}{\partial \bm{\theta}} Q_\Omega(s',o^{1:N-1}) \\ +
(1-\beta^{N-1}(s',o^{1:N-1})) \frac{\partial Q_\Omega(s',o^{1:N-1})}{\partial \bm{\theta}} \\ +
 \frac{ \partial V_\Omega(s')}{\partial \bm{\theta}} \prod_{j=N-1}^{1} \beta^j(s',o^{1:j}) \\ + V_\Omega(s') \sum_{k=N-1}^1 \frac{\partial \beta^k(s',o^{1:k})}{\partial \bm{\theta}}  \prod_{\substack{j=N-1 \\ j \neq k}}^{1} \beta^j(s',o^{1:j}) \\
- \sum_{i=1}^{N - 2}\frac{ \partial \beta^i(s',o^{1:i})}{\partial \bm{\theta}} Q_\Omega(s',o^{1:i}) \prod_{k=N-1}^{i+1} \beta^k(s',o^{1:k}) \\
+ \sum_{i=1}^{N - 2}(1-\beta^i(s',o^{1:i})) \frac{\partial Q_\Omega(s',o^{1:i})}{\partial \bm{\theta}} \prod_{k=N-1}^{i+1} \beta^k(s',o^{1:k}) \\ 
+ \sum_{i=1}^{N - 2}(1-\beta^i(s',o^{1:i})) Q_\Omega(s',o^{1:i}) \bigg( \\ \sum_{j=i+1}^{N-1} \frac{ \partial \beta^j(s',o^{1:j})}{\partial \bm{\theta}} \prod_{\substack{k=N-1 \\ k \neq j}}^{i+1} \beta^k(s',o^{1:k}) \bigg).
\end{split}
\end{equation}

We can define the value function $V_\Omega(s')$ in terms of the option-value function using the policy over options at each layer:

\begin{equation} 
V_\Omega(s') = \sum_{o'^{1:N-1}} \prod_{i=1}^{N-1} \pi^i(o'^i|s',o'^{1:i-1}) Q_\Omega(s',o'^{1:N-1}).
\end{equation}

Then we can then take the gradient to obtain:

\begin{equation}  \label{eq33}
\begin{split}
\frac{\partial V_\Omega(s')}{\partial \bm{\theta}} = \sum_{o'^{1:N-1}} \bigg( \prod_{i=1}^{N-1} \pi^i(o'^i|s',o'^{1:i-1}) \frac{\partial Q_\Omega(s',o'^{1:N-1})}{\partial \bm{\theta}} \\
+ \sum_{j=1}^{N-1} \frac{ \partial \pi^j(o'^j|s',o'^{1:j-1})}{\partial \bm{\theta}} \prod_{\substack{i=1 \\ i \neq j}}^{N-1} \pi^i(o'^i|s',o'^{1:i-1}) Q_\Omega(s',o'^{1:N-1}) \bigg).
\end{split}
\end{equation}

Likewise, we can define the option-value function $Q_\Omega(s,o^{1:i})$ by integrating out the option-value function using the policy over options at each layer:

\begin{equation} 
Q_\Omega(s,o^{1:i}) = \sum_{o'^{i+1:N-1}} \prod_{i=i+1}^{N-1} \pi^i(o^i|s,o^{1:i-1}) Q_\Omega(s,o^{1:N-1}).
\end{equation}

Then we can then take the gradient to obtain:

\begin{equation}   \label{eq35}
\begin{split}
\frac{\partial Q_\Omega(s,o^{1:i})}{\partial \bm{\theta}} = \\ \sum_{o^{i+1:N-1}} \bigg( \prod_{j=i+1}^{N-1} \pi^j(o^j|s,o^{1:j-1}) \frac{\partial Q_\Omega(s,o^{1:N-1})}{\partial \bm{\theta}}
 + \\ \sum_{j=i+1}^{N-1} \frac{ \partial \pi^j(o^j|s,o^{1:j-1})}{\partial \bm{\theta}} \prod_{\substack{k=i+1 \\ k \neq j}}^{N-1} \pi^k(o^k|s,o^{1:k-1}) Q_\Omega(s,o^{1:N-1}) \bigg).
\end{split}
\end{equation}

We can now simplify our original expression in equation \ref{eq31} in terms of the derivative of the option-value function. We start by keeping the first two terms of equation \ref{eq31}. For the third term, we substitute in equation \ref{eq33}. We keep the rest of the terms unaltered with the exception of the sixth term, for which we substitute in equation \ref{eq35} evaluated at $s'$. 

\begin{footnotesize}
\begin{equation} 
\begin{split}
\frac{\partial U(s',o^{1:N-1})}{\partial \bm{\theta}} = -\frac{ \partial \beta^{N-1}(s',o^{1:N-1})}{\partial \bm{\theta}} Q_\Omega(s',o^{1:N-1}) \\ +
(1-\beta^{N-1}(s',o^{1:N-1})) \frac{\partial Q_\Omega(s',o^{1:N-1})}{\partial \bm{\theta}} \\+
 \sum_{o'^{1:N-1}} \bigg( \prod_{i=1}^{N-1} \pi^i(o'^i|s',o'^{1:i-1}) \frac{\partial Q_\Omega(s',o'^{1:N-1})}{\partial \bm{\theta}} \\
+ \sum_{j=1}^{N-1} \frac{ \partial \pi^j(o'^j|s',o'^{1:j-1})}{\partial \bm{\theta}} \\ \prod_{\substack{i=1 \\ i \neq j}}^{N-1} \pi^i(o'^i|s',o'^{1:i-1}) Q_\Omega(s',o'^{1:N-1}) \bigg) \prod_{j=N-1}^{1} \beta^j(s',o^{1:j}) \\ + V_\Omega(s') \sum_{k=N-1}^1 \frac{\partial \beta^k(s',o^{1:k})}{\partial \bm{\theta}}  \prod_{\substack{j=N-1 \\ j \neq k}}^{1} \beta^j(s',o^{1:j}) \\
- \sum_{i=1}^{N - 2}\frac{ \partial \beta^i(s',o^{1:i})}{\partial \bm{\theta}} Q_\Omega(s',o^{1:i}) \prod_{k=N-1}^{i+1} \beta^k(s',o^{1:k}) \\
+ \sum_{i=1}^{N - 2}(1-\beta^i(s',o^{1:i})) \sum_{o^{i+1:N-1}} \bigg( \\ \prod_{j=i+1}^{N-1} \pi^j(o'^j|s',o'^{1:j-1}) \frac{\partial Q_\Omega(s',o^{1:N-1})}{\partial \bm{\theta}} \\
+ \sum_{j=i+1}^{N-1} \frac{ \partial \pi^j(o'^j|s',o'^{1:j-1})}{\partial \bm{\theta}} \\ \prod_{\substack{k=i+1 \\ k \neq j}}^{N-1} \pi^k(o'^k|s',o'^{1:k-1}) Q_\Omega(s',o^{1:N-1}) \bigg) \prod_{k=N-1}^{i+1} \beta^k(s',o^{1:k}) \\ 
+ \sum_{i=1}^{N - 2}(1-\beta^i(s',o^{1:i})) Q_\Omega(s',o^{1:i}) \bigg( \\ \sum_{j=i+1}^{N-1} \frac{ \partial \beta^j(s',o^{1:j})}{\partial \bm{\theta}} \prod_{\substack{k=N-1 \\ k \neq j}}^{i+1} \beta^k(s',o^{1:k}) \bigg).
\end{split}
\end{equation}
\end{footnotesize}

We can then reformulate this expression even further by collecting terms that depend on the derivative of the option-value function. These terms include the original second term, the first part of the expanded third term and the first part of the expanded sixth term. 

\begin{footnotesize}
\begin{equation} 
\begin{split}
 \frac{\partial U(s',o^{1:N-1})}{\partial \bm{\theta}} = -\frac{ \partial \beta^{N-1}(s',o^{1:N-1})}{\partial \bm{\theta}} Q_\Omega(s',o^{1:N-1}) \\ +
\bigg( (1-\beta^{N-1}(s',o^{1:N-1})) \textbf{1}_{o'^{1:N-1}=o^{1:N-1}} \\ +
\prod_{j=N-1}^{1} \beta^j(s',o^{1:j})  \pi^j(o'^j|s_{t+1},o'^{1:j-1}) 
\\+ \sum_{i=1}^{N - 2}(1-\beta^i(s',o^{1:i})) \textbf{1}_{o'^{1:i}=o^{1:i}} \bigg[ \\ \prod_{k=N-1}^{i+1} \beta^k(s',o^{1:k}) \pi^k(o'^k|s',o'^{1:k-1}) \bigg]  \bigg) \frac{\partial Q_\Omega(s',o'^{1:N-1})}{\partial \bm{\theta}}
\\+ \prod_{j=N-1}^{1} \beta^j(s',o^{1:j}) \sum_{o'^{1:N-1}} \sum_{j=1}^{N-1} \bigg[ \\ \frac{ \partial \pi^j(o'^j|s',o'^{j-1})}{\partial \bm{\theta}} \prod_{\substack{i=1 \\ i \neq j}}^{N-1} \pi^i(o'^i|s',o'^{i-1}) Q_\Omega(s',o'^{1:N-1}) \bigg] \\
\\+ \sum_{i=1}^{N - 2}(1-\beta^i(s',o^{1:i})) \prod_{k=N-1}^{i+1} \beta^k(s',o^{1:k}) \sum_{o^{i+1:N-1}}  \sum_{j=i+1}^{N-1} \bigg[\\ \frac{ \partial \pi^j(o'^j|s',o'^{1:j-1})}{\partial \bm{\theta}} \\ \prod_{\substack{k=i+1 \\ k \neq j}}^{N-1} \pi^k(o'^k|s',o'^{1:k-1}) Q_\Omega(s',o^{1:N-1}) \bigg]\\
+ V_\Omega(s') \sum_{k=N-1}^1 \frac{\partial \beta^k(s',o^{1:k})}{\partial \bm{\theta}}  \prod_{\substack{j=N-1 \\ j \neq k}}^{1} \beta^j(s',o^{1:j}) 
\\ - \sum_{i=1}^{N - 2}\frac{ \partial \beta^i(s',o^{1:i})}{\partial \bm{\theta}} Q_\Omega(s',o^{1:i}) \prod_{k=N-1}^{i+1}  \beta^k(s',o^{1:k})
\\+ \sum_{i=1}^{N - 2}(1-\beta^i(s',o^{1:i})) Q_\Omega(s',o^{1:i}) \bigg( \sum_{j=i+1}^{N-1} \frac{ \partial \beta^j(s',o^{1:j})}{\partial \bm{\theta}} \\ \prod_{\substack{k=N-1 \\ k \neq j}}^{i+1} \beta^k(s',o^{1:k}) \bigg).
\end{split}
\end{equation}
\end{footnotesize}

We now substitute this last expression into equation \ref{QOmega}, which allows us to also factor out equation \ref{P1HOC}:

\begin{equation} 
\begin{split}
\frac{\partial Q_\Omega(s,o^{1:N-1})}{\partial \bm{\theta}} 
= \sum_{a} \frac{ \partial \pi^N(a|s,o^{1:N-1})}{\partial \bm{\theta}} Q_U(s,o^{1:N-1},a) \\
+  \sum_{s'} \sum_{o'^{1:N-1}} P^{(1)}_\gamma(s',o'^{1:N-1}|s,o^{1:N-1}) \frac{\partial Q_\Omega(s',o'^{1:N-1})}{\partial \bm{\theta}} \\
+ \sum_{a} \pi^N(a|s,o^{1:N-1}) \gamma \sum_{s'} P (s'|s,a) \bigg[ 
\\-\frac{ \partial \beta^{N-1}(s',o^{1:N-1})}{\partial \bm{\theta}} Q_\Omega(s',o^{1:N-1}) \\
+ \prod_{j=N-1}^{1} \beta^j(s',o^{1:j}) \sum_{o'^{1:N-1}} \sum_{j=1}^{N-1} \bigg[ \\ \frac{ \partial \pi^j(o'^j|s',o'^{j-1})}{\partial \bm{\theta}} \prod_{\substack{i=1 \\ i \neq j}}^{N-1} \pi^i(o'^i|s',o'^{i-1}) Q_\Omega(s',o'^{1:N-1}) \bigg] \\
+ \sum_{i=1}^{N - 2}(1-\beta^i(s',o^{1:i})) \prod_{k=N-1}^{i+1} \beta^k(s',o^{1:k}) \sum_{o^{i+1:N-1}} \sum_{j=i+1}^{N-1} \big[ \\ \frac{ \partial \pi^j(o'^j|s',o'^{1:j-1})}{\partial \bm{\theta}} \prod_{\substack{k=i+1 \\ k \neq j}}^{N-1} \pi^k(o'^k|s',o'^{1:k-1}) Q_\Omega(s',o'^{1:N-1}) \big]\\
+ V_\Omega(s') \sum_{k=N-1}^1 \frac{\partial \beta^k(s',o^{1:k})}{\partial \bm{\theta}}  \prod_{\substack{j=N-1 \\ j \neq k}}^{1} \beta^j(s',o^{1:j})
\\- \sum_{i=1}^{N - 2}\frac{ \partial \beta^i(s',o^{1:i})}{\partial \bm{\theta}} Q_\Omega(s',o^{1:i}) \prod_{k=N-1}^{i+1} \beta^k(s',o^{1:k}) \\
\\+ \sum_{i=1}^{N - 2}(1-\beta^i(s',o^{1:i})) Q_\Omega(s',o^{1:i}) \bigg( \\ \sum_{j=i+1}^{N-1} \frac{ \partial \beta^j(s',o^{1:j})}{\partial \bm{\theta}} \prod_{\substack{k=N-1 \\ k \neq j}}^{i+1} \beta^k(s',o^{1:k}) \bigg)
\bigg],
\end{split}
\end{equation}

We can then reorganize the terms of this expression. We keep the first two terms the same. The third term is now the old sixth term. The new fourth term is a combination of the old seventh term and third term. The fifth term is now the old eighth term. The sixth term is still the same as before and the seventh term is the old fourth term. \footnote{Here we define that $\beta(s',o^{1:N})=1$.}

\begin{equation} 
\begin{split}
\frac{\partial Q_\Omega(s,o^{1:N-1})}{\partial \bm{\theta}} 
= \sum_{a} \frac{ \partial \pi^N(a|s,o^{1:N-1})}{\partial \bm{\theta}} Q_U(s,o^{1:N-1},a) \\
+  \sum_{s'} \sum_{o'^{1:N-1}} P^{(1)}_\gamma(s',o'^{1:N-1}|s,o^{1:N-1}) \frac{\partial Q_\Omega(s',o'^{1:N-1})}{\partial \bm{\theta}}
\\+ \sum_{a} \pi^N(a|s,o^{1:N-1}) \gamma \sum_{s'} P (s'|s,a) \bigg[ \\
V_\Omega(s') \sum_{k=N-1}^1 \frac{\partial \beta^k(s',o^{1:k})}{\partial \bm{\theta}}  \prod_{\substack{j=N-1 \\ j \neq k}}^{1} \beta^j(s',o^{1:j}) \\
- \sum_{i=1}^{N - 1}\frac{ \partial \beta^i(s',o^{1:i})}{\partial \bm{\theta}} Q_\Omega(s',o^{1:i}) \prod_{k=N-1}^{i+1} \beta^k(s',o^{1:k}) \\
+ \sum_{i=1}^{N - 2}(1-\beta^i(s',o^{1:i})) Q_\Omega(s',o^{1:i}) \bigg( \\ \sum_{j=i+1}^{N-1} \frac{ \partial \beta^j(s',o^{1:j})}{\partial \bm{\theta}} \prod_{\substack{k=N-1 \\ k \neq j}}^{i+1} \beta^k(s',o^{1:k}) \bigg) \\
+ \sum_{i=1}^{N - 2}(1-\beta^i(s',o^{1:i})) \prod_{k=N-1}^{i+1} \beta^k(s',o^{1:k}) \sum_{o^{i+1:N-1}} \sum_{j=i+1}^{N-1} \big[ \\ \frac{ \partial \pi^j(o'^j|s',o'^{1:j-1})}{\partial \bm{\theta}} \prod_{\substack{k=i+1 \\ k \neq j}}^{N-1} \pi^k(o'^k|s',o'^{1:k-1}) Q_\Omega(s',o'^{1:N-1}) \big]\\
\\+ \prod_{j=N-1}^{1} \beta^j(s',o^{1:j}) \sum_{o'^{1:N-1}} \sum_{j=1}^{N-1} \bigg(\\ \frac{ \partial \pi^j(o'^j|s',o'^{j-1})}{\partial \bm{\theta}} \prod_{\substack{i=1 \\ i \neq j}}^{N-1} \pi^i(o'^i|s',o'^{i-1}) Q_\Omega(s',o'^{1:N-1}) \bigg) 
\bigg],
\end{split}
\end{equation}

As in \citet{HOC} we can further condense our expression by noting that the generalized advantage function over a hierarchical set of options can be defined as $A_\Omega(s',o^{1:\ell}) = Q_\Omega(s',o^{1:\ell}) - V_\Omega(s') [\prod_{j=\ell-1}^{1} \beta^j(s',o^{1:j})]  - \sum_{i=1}^{\ell - 1}(1-\beta^i(s',o^{1:i})) Q_\Omega(s',o^{1:i}) [\prod_{k=i+1}^{\ell - 1} \beta^k(s',o^{1:k})]$. We keep the first two terms in the last expression unchanged. Then we note that the third, fourth, and fifth terms in the last equation can be combined using the generalized advantage function notation. This can be seen by renaming $k$ in the third term to $\ell$, and renaming $i$ in the fourth and fifth terms to $\ell$, which allows us to bring out the same sum over all terms. Finally, we keep the sixth and seventh terms unchanged from the last equation. 

\begin{equation} 
\begin{split}
\frac{\partial Q_\Omega(s,o^{1:N-1})}{\partial \bm{\theta}} 
= \sum_{a} \frac{ \partial \pi^N(a|s,o^{1:N-1})}{\partial \bm{\theta}} Q_U(s,o^{1:N-1},a) \\
+  \sum_{s'} \sum_{o'^{1:N-1}} P^{(1)}_\gamma(s',o'^{1:N-1}|s,o^{1:N-1}) \frac{\partial Q_\Omega(s',o'^{1:N-1})}{\partial \bm{\theta}} \\
+ \sum_{a} \pi^N(a|s,o^{1:N-1}) \gamma \sum_{s'} P (s'|s,a) \bigg[ \\ 
 \sum_{\ell=1}^{N-1} \frac{\partial \beta^\ell(s',o^{1:\ell})}{\partial \bm{\theta}} A_\Omega(s',o^{1:\ell}) [ \prod_{k=N-1}^{\ell+1} \beta^k(s',o^{1:k}) ] \\
+ \sum_{i=1}^{N - 2}(1-\beta^i(s',o^{1:i})) \prod_{k=N-1}^{i+1} \beta^k(s',o^{1:k}) \sum_{o^{i+1:N-1}} \sum_{j=i+1}^{N-1} \big[\\ \frac{ \partial \pi^j(o'^j|s',o'^{1:j-1})}{\partial \bm{\theta}} \prod_{\substack{k=i+1 \\ k \neq j}}^{N-1} \pi^k(o'^k|s',o'^{1:k-1}) Q_\Omega(s',o'^{1:N-1}) \big]\\
\\+ \prod_{j=N-1}^{1} \beta^j(s',o^{1:j}) \sum_{o'^{1:N-1}} \sum_{j=1}^{N-1} \bigg( \\ \frac{ \partial \pi^j(o'^j|s',o'^{1:j-1})}{\partial \bm{\theta}} \prod_{\substack{i=1 \\ i \neq j}}^{N-1} \pi^i(o'^i|s',o'^{1:i-1}) Q_\Omega(s',o'^{1:N-1}) \bigg) 
\bigg].
\end{split}
\end{equation}

We can also condense the terms related to the gradient of $\pi^\ell$ at each level of the option hierarchy $\ell$ by using the higher level option policies to integrate out the option-value function. We can then define the probability while at the next state and terminating the options for the current state that the agent arrives at a particular set of next option selections i.e. $P_{\pi,\beta}(o'^{1:\ell-1}|s',o^{1:\ell-1})$.   Where we define $P_{\pi,\beta}(o'^{1:\ell-1}|s',o^{1:\ell-1}) = \sum_{i=1}^{\ell - 1}(1-\beta^i(s',o^{1:i})) \prod_{k=\ell-1}^{i+1} \beta^k(s',o^{1:k}) \pi^k(o'^k|s',o'^{1:k-1})
+ \prod_{j=\ell-1}^{1} \beta^j(s',o^{1:j}) \pi^j(o'^j|s',o'^{1:j-1})$:

\begin{equation} 
\begin{split}
\frac{\partial Q_\Omega(s,o^{1:N-1})}{\partial \bm{\theta}} 
= \sum_{a} \frac{ \partial \pi^N(a|s,o^{1:N-1})}{\partial \bm{\theta}} Q_U(s,o^{1:N-1},a) \\
+  \sum_{s'} \sum_{o'^{1:N-1}} P^{(1)}_\gamma(s',o'^{1:N-1}|s,o^{1:N-1}) \frac{\partial Q_\Omega(s',o'^{1:N-1})}{\partial \bm{\theta}} \\
+ \sum_{a} \pi^N(a|s,o^{1:N-1}) \gamma \sum_{s'} P (s'|s,a) \bigg[ \\ 
 \sum_{\ell=1}^{N-1} \frac{\partial \beta^\ell(s',o^{1:\ell})}{\partial \bm{\theta}} A_\Omega(s',o^{1:\ell}) [ \prod_{k=N-1}^{\ell+1} \beta^k(s',o^{1:k}) ] \\
+ \sum_{o'^{1:N-1}} \sum_{\ell=1}^{N-1} \frac{\partial \pi^\ell(o'^\ell|s',o'^{1:\ell-1})}{\partial \bm{\theta}} Q_\Omega(s',o'^{1:\ell}) [ \\ \prod_{k=N-1}^{\ell} \beta^k(s',o^{1:k})] P_{\pi,\beta}(o'^{1:\ell-1}|s',o^{1:\ell-1}) \bigg] .
\end{split}
\end{equation}

We can simplify this expression using the fact that $P(s'|s,o^{1:N-1}) = \sum_a \pi(a|s,o^{1:N-1})P(s'|s,a)$. We can then also rearrange this expression for simplicity by multiplying the first term by $\sum_{s'} P(s'|s,o^{1:N-1})=1$ as the term does not depend on $s'$. 

\begin{equation} 
\begin{split}
\frac{\partial Q_\Omega(s,o^{1:N-1})}{\partial \bm{\theta}} 
= \sum_{a} \frac{ \partial \pi^N(a|s,o^{1:N-1})}{\partial \bm{\theta}} Q_U(s,o^{1:N-1},a) \\
+  \sum_{s'} \sum_{o'^{1:N-1}} P^{(1)}_\gamma(s',o'^{1:N-1}|s,o^{1:N-1}) \frac{\partial Q_\Omega(s',o'^{1:N-1})}{\partial \bm{\theta}}
\\+ \gamma \sum_{s'} P(s'|s,o^{1:N-1}) \bigg[ \\ 
-\sum_{o^{1:N-1}} \frac{\partial \beta^\ell(s',o^{1:\ell})}{\partial \bm{\theta}} A_\Omega(s',o^{1:\ell}) [ \prod_{k=N-1}^{\ell+1} \beta^k(s',o^{1:k}) ] \\
+ \sum_{o'^{1:N-1}} \sum_{\ell=1}^{N-1} \frac{\partial \pi^\ell(o'^\ell|s',o'^{1:\ell-1})}{\partial \bm{\theta}} Q_\Omega(s',o'^{1:\ell}) [ \\ \prod_{k=N-1}^{\ell} \beta^k(s',o^{1:k}) ] \\
= \sum_{s'} \sum_{o'^{1:N-1}} P^{(1)}_\gamma(s',o'^{1:N-1}|s,o^{1:N-1}) \frac{\partial Q_\Omega(s',o'^{1:N-1})}{\partial \bm{\theta}} \\ + \sum_{s'} P(s'|s,o^{1:N-1}) \bigg[ \\ \sum_{a} \frac{ \partial \pi^N(a|s,o^{1:N-1})}{\partial \bm{\theta}} Q_U(s,o^{1:N-1},a) \\
- \gamma  \sum_{\ell=1}^{N-1} \frac{\partial \beta^\ell(s',o^{1:\ell})}{\partial \bm{\theta}} A_\Omega(s',o^{1:\ell}) [ \prod_{k=N-1}^{\ell+1} \beta^k(s',o^{1:k}) ] \\
+ \gamma  \sum_{o'^{1:N-1}} \sum_{\ell=1}^{N-1} \frac{\partial \pi^\ell(o'^\ell|s',o'^{1:\ell-1})}{\partial \bm{\theta}} Q_\Omega(s',o'^{1:\ell}) [\\ \prod_{k=N-1}^{\ell} \beta^k(s',o^{1:k}) ]  P_{\pi,\beta}(o'^{1:\ell-1}|s',o^{1:\ell-1})
\bigg].
\end{split}
\end{equation}

Using the structure of the augmented process from the previous section and the fact that $P_\gamma^{(k)}(s',o'^{1:N-1},s''|s,o^{1:N-1}) = P(s''|s',o'^{1:N-1}) P_\gamma^{(k)}(s',o'^{1:N-1}|s,o^{1:N-1})$
we finally obtain:

\begin{equation}
\begin{split}
\frac{\partial Q_\Omega(s,o^{1:N-1})}{\partial \bm{\theta}} = \\ \sum_{s',o'^{1:N-1},s''} \sum_{k=0}^\infty P^{(k)}_\gamma(s',o'^{1:N-1},s''|s,o^{1:N-1})  \bigg(\\ \sum_{a'} \frac{\partial \pi(a'|s',o'^{1:N-1})}{\partial \bm{\theta}} Q_U(s',o'^{1:N-1},a') \\ + \gamma \sum_{o''^{1:N-1}}  \sum_{\ell=1}^{N-1} \frac{\partial \pi^\ell(o''^\ell|s'',o''^{1:\ell-1})}{\partial \bm{\theta}} Q_\Omega(s'',o''^{1:\ell}) \bigg[ \\ \prod_{k=N-1}^{\ell} \beta^k(s'',o'^{1:k})  P_{\pi,\beta}(o''^{1:\ell-1}|s'',o'^{1:\ell-1}) \bigg] \\ - \gamma \sum_{\ell=1}^{N-1} \frac{\partial \beta^\ell(s'',o'^{1:\ell})}{\partial \bm{\theta}} A_\Omega(s'',o'^{1:\ell}) \prod_{k=N-1}^{\ell+1} \beta^k(s'',o'^{1:k}) \bigg).
\end{split}
\end{equation}

Therefore the gradient of the expected discounted return with respect to $\bm{\theta}$ is:

\begin{equation}
\begin{split}
\frac{\partial Q_\Omega(s_0,o^{1:N-1}_0)}{\partial \bm{\theta}} = \\ \sum_{s,o^{1:N-1},s'} \sum_{k=0}^\infty P^{(k)}_\gamma(s,o^{1:N-1},s'|s_0,o_0^{1:N-1}) \\ \bigg( \sum_{a}\frac{\partial \pi(a|s,o^{1:N-1})}{\partial \bm{\theta}} Q_U(s,o^{1:N-1},a) \\ + \gamma \sum_{o'^{1:N-1}} \sum_{\ell=1}^{N-1} \frac{\partial \pi^\ell(o'^\ell|s',o'^{1:\ell-1})}{\partial \bm{\theta}} Q_\Omega(s',o'^{1:\ell}) \bigg[ \\ \prod_{k=N-1}^{\ell} \beta^k(s',o^{1:k})  P_{\pi,\beta}(o'^{1:\ell-1}|s',o^{1:\ell-1}) \bigg] \\ - \gamma \sum_{\ell=1}^{N-1} \frac{\partial \beta^\ell(s',o^{1:\ell})}{\partial \bm{\theta}} A_\Omega(s',o^{1:\ell}) \prod_{k=N-1}^{\ell+1} \beta^k(s',o^{1:k}) \bigg) \\
= \sum_{s,o^{1:N-1},s'}  \mu_\Omega(s,o^{1:N-1},s'|s_0,o_0^{1:N-1})  \bigg( \\ \sum_{a} \frac{\partial \pi(a|s,o^{1:N-1})}{\partial \bm{\theta}} Q_U(s,o^{1:N-1},a) \\ + \gamma \sum_{o'^{1:N-1}} \sum_{\ell=1}^{N-1} \frac{\partial \pi^\ell(o'^\ell|s',o'^{1:\ell-1})}{\partial \bm{\theta}} Q_\Omega(s',o'^{1:\ell}) \bigg[ \\ \prod_{k=N-1}^{\ell} \beta^k(s',o^{1:k})  P_{\pi,\beta}(o'^{1:\ell-1}|s',o^{1:\ell-1}) \bigg] \\ - \gamma \sum_{\ell=1}^{N-1} \frac{\partial \beta^\ell(s',o^{1:\ell})}{\partial \bm{\theta}} A_\Omega(s',o^{1:\ell}) \prod_{k=N-1}^{\ell+1} \beta^k(s',o^{1:k}) \bigg).
\end{split}
\end{equation}

\section{Appendix C: Deriving Lemma 1 and Theorem 1 by Extending Coagent Networks}

The recent work of \citet{kostas2019reinforcement} first suggested that option-critic is a special incarnation of a very general reinforcement learning framework called Coagent Networks. In their paper they revealed that there is a general purpose policy gradient theorem for all Coagent Networks and that the policy gradient theorem of option-critic can be derived by mapping the process flow of option-critic in a way that can be described as a graph of \textit{coagents} i.e. generic computation nodes. In this appendix we would like to show that with our more general assumptions about parameter sharing, leveraging the coagent policy gradient theorem does indeed yield the policy gradient theorems that we derived in the earlier sections of the appendix. 

\subsection{Background on Coagent Networks}

``Coagent Networks'' (CN) were introduced in \cite{coagents}, and model the mapping of any kind of policy to a graph of computation nodes, so-called \textit{coagents}. Utilizing the coagent policy gradient theorems in \cite{thomas2011policy,kostas2019reinforcement}, this mapping can then be used to dramatically simplify derivations of policy gradient theorems for complex policies.

We define a coagent $i$ as a generic computation node that takes in some kind of input $x \in X_i$ and computes some kind of output $u \in U_i$ using the function  $u=\kappa_i(x)$. $J$ is the score function, $d$ is the distribution over start states, and $Q$ is the state-action value function. The following equation describes the local policy gradient (with respect to some parameters $\theta$) for each independent computation node, where ``independent'' means that the respective nodes are d-separable:
\begin{align}
	\frac{\partial J(\theta)}{\partial \theta_i} &= \sum_{x\in X_i}d_i^\pi \sum_{u \in U_i} \frac{\partial \kappa_i (x, u)}{\partial \theta_i} Q_{U_i}(x, u). \label{eq:capg}
\end{align}
This equation states that the local policy gradient for one coagent can be derived by marginalizing over its input ($X_i$) and output ($U_i$) spaces. A Coagent Network can be defined by connecting a graph of coagents. It has been shown that computing this local policy gradient theorem for each coagent in the graph in sufficient to compute the policy gradient for the full graph. It is important to note that this fact was only proven by \cite{kostas2019reinforcement} for the two extreme cases of totally independent co-agents and repeated use of the same co-agent. However, this also holds for co-agents with partial weight sharing. We leave the formal proof to future work.

\subsection{Using Coagent Networks to Derive Lemma 1}
\begin{figure}[h]
	\centering
	\begin{tikzpicture}
		\node[multiplexer]
		(mul) at (3,0.8) {};
		
		\node (piO) at (2,1.6) {$\pi_\Omega$};
		\node (st) at (0.,1.6) {$s_t$};
		\node (omegat1) at (0.,0.8) {$o_{t-1}$};
		\node (beta) at (2,0) {$\beta_o$};
		\node (pio) at (4,1.6) {$\pi_o$};
		\node (omegat) at (5.,0.8) {$o_t$};
		\node (at) at (5.,1.6) {$a_t$};

		\node (sto) at (1,1.6) {};
		\node (p2) at (1,0.8) {};
		\node (ot1beta) at (2,0.8) {};
		\node (mulpio) at (4,0.8) {};
		
		\draw[->] (st) to (piO);
		\draw[->] (pio) to (at);
		\draw[->] (mul.top side) to (omegat);
		\draw[->] (omegat1) to (omegat1);
		\draw[->] (beta) to (3,0) to (mul.south);
		\draw[->] (1,1.6) to ($(sto)+(0,0.5)$) to ($(pio)+(0,0.5)$) to (pio);
		\draw[->] (1,1.6) to (1.,0) to (beta);
		\draw[->] (omegat1) to (p2) to ($(p2)+(1,0)$) |- (mul.south west);
		\draw[->] ($(p2)+(1,0)$) to (beta);
		\draw[->] (4,0.8) to (pio);
		\draw[->] (piO) to ($(piO)+(0.6,0)$) |- (mul.north west);
	\end{tikzpicture}
	\caption{The coagent network for the option-critic architecture, as introduced in \cite{kostas2019reinforcement}.}
	\label{fig:coagent-ocpg}
\end{figure}
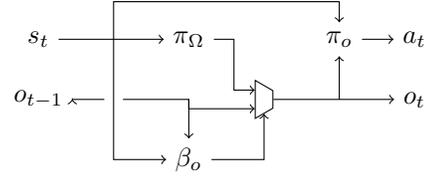
This derivation consists of two parts. The first is to find the graph that adequately models the policy. For the option-critic architecture, \cite{kostas2019reinforcement} already defines a graph mapping the option-critic architecture, depicted in Figure \ref{fig:coagent-ocpg}. However, they also assume independence of parameters across architecture components, deriving the original theorem in \cite{OC}.

The trick is that we can use d-separation on the dependency graph for the different components of the coagent network modeling the option-critic framework. When we do so, we get two separate subnetworks. The first is still the independent option-policy node (computing $\pi(s,o)$). Plugging this into equation \eqref{eq:capg} (with $\kappa=\pi$) immediately yields

\begin{equation}
\begin{split}
\frac{\partial J_1}{\partial \bm{\theta}} = \sum_{s, o, s'} \mu_\Omega (s, o, s'| s_0, o_0) \sum_a \frac{\partial \pi(a|s,o)}{\partial \bm{\theta}} Q_U(s, o, a) 
\end{split}
\end{equation}
	
as $\mu$ is our initial distribution, $s,o,s'$ our inputspace and $a$ our output space, both of which we marginalize. More difficult are $\beta$ and $\pi_\Omega$, as these cannot be separated by d-separation. Consequently, our first step is to formulate the joint function $\kappa$ computed by $\beta$ and $\pi_\Omega$, which is denoted $\pi_i$ in equation \eqref{eq:capg}. By convention, $\beta$ and $\pi_\Omega$ are only evaluated at the next state, so 
we need to consider state $s'$ instead of $s$ and option $o'$ for the policy over options:

\begin{equation}
\begin{split}
\frac{\partial J_1}{\partial \bm{\theta}} = \sum_{s, o, s'} \mu_\Omega (s, o, s'| s_0, o_0) \sum_a \frac{\partial \pi(a|s,o)}{\partial \bm{\theta}} Q_U(s, o, a) 
\end{split}
\end{equation}

\begin{equation}
\begin{split}
\kappa_{\pi,\beta}(o'|s, o) = \beta(s', o)\pi_\Omega(o', s') + (1-\beta(s', o)) \textbf{1}_{o' = o}
\end{split}
\end{equation}

Before combining this with equation \ref{eq:capg}, we need to consider that \eqref{eq:capg} determines the value of taking an action and evaluating it immediately. In line with computing the value at state $s'$, we need to discount their value:

\begin{equation}
\begin{split}
\frac{\partial J_2}{\partial \bm{\theta}} = \gamma\sum_{s, o, s'} \mu_\Omega (s, o, s'| s_0, o_0) \sum_{o'} \frac{\partial }{\partial \bm{\theta}} \bigg( \beta(s', o)\pi_\Omega(o', s') \\+ (1-\beta(s', o))  \textbf{1}_{o' = o} \bigg)Q_\Omega(s', o)
\end{split}
\end{equation}

The output space is the set of options, but the input spaces are the same. This is equivalent to (dropping the arguments of $\mu$ for readability):

\begin{equation}
\begin{split}
\frac{\partial J_2}{\partial \bm{\theta}} = \gamma\sum_{s, o, s'} \mu_\Omega \sum_{o'} \frac{\partial }{\partial \bm{\theta}} \bigg(\beta(s', o)\pi_\Omega(o', s') Q_\Omega(s', o') \\+ (1-\beta)Q_\Omega(s', o) \bigg)\\
=\gamma\sum_{s, o, s'} \mu_\Omega \sum_{o'} \bigg( \beta(s', o)\frac{\partial }{\partial \bm{\theta}}\pi_\Omega(o', s') Q_\Omega(s', o') + \\ \frac{\partial }{\partial \bm{\theta}}\beta(s', o)\underbrace{\pi_\Omega(o', s') Q_\Omega(s', o')}_{=V_\Omega(s')} - \frac{\partial }{\partial \bm{\theta}}\beta Q_\Omega(s', o) \bigg)\\
= \gamma\sum_{s, o, s'} \mu_\Omega \sum_{o'} \bigg( \\ \beta(s', o)\frac{\partial }{\partial \bm{\theta}}\pi_\Omega(o', s') Q_\Omega(s, o') - \frac{\partial }{\partial \bm{\theta}}\beta A_\Omega(s', o) \bigg)
\end{split}
\end{equation}

Finally we note that the derivative of the total network objective for all parameters is the same as the added derivative with respect to the two subnetworks:

\begin{equation} 
\begin{split}
\frac{\partial J}{\partial \bm{\theta}} = \frac{\partial J_1}{\partial \bm{\theta}} + \frac{\partial J_2}{\partial \bm{\theta}}
		= \sum_{s,o,s'} \mu_\Omega(s,o,s'|s_0,o_0) \bigg( \\ \sum_{a} \frac{\partial \pi(a|s,o)}{\partial \bm{\theta}} Q_U(s,o,a) + \\ \sum_{o'} \gamma \beta(s',o) \frac{\partial \pi_\Omega(o'|s')}{\partial \bm{\theta}} Q_\Omega(s',o') - \gamma \frac{\partial \beta(s',o)}{\partial \bm{\theta}}A_\Omega(s',o) \bigg) 
\end{split}
\end{equation}

We have derived the update rule from Lemma 1. We now will follow similar logic for Theorem 1. 

\subsection{Using Coagent Networks to Derive Theorem 1}

\begin{figure*}[h]
	\centering
	\begin{tikzpicture}
		(mul) at (3,0.8) {};
		
		\node (two) at (-5,1.6) {$\beta^{N-1}$};
		\node (one) at (-7,1.6) {$s_t,o_{t-1}^{1:N-1}$};
		\node (three) at (-4,1.6) {...};
		\node (four) at (-3,1.6) {$\beta^{1}$};
		\node (five) at (-2,1.6) {$\pi^1$};
		\node (six) at (-1,1.6) {...};
		\node (seven) at (0,1.6) {$\pi^N$};
		\node (eight) at (1.5,1.6) {$s_t,o_{t}^{1:N}$};
		\node (nine) at (3.0,1.6) {$env$};
		\node (ten) at (4.5,1.6) {$s_{t+1},o_{t}^{1:N}$};
		\node (eleven) at (6,1.6) {...};

		
		\draw[->] (one) to (two);
		\draw[->] (two) to (three);
		\draw[->] (three) to (four);
		\draw[->] (four) to (five);
		\draw[->] (five) to (six);
		\draw[->] (six) to (seven);
		\draw[->] (seven) to (eight);
		\draw[->] (eight) to (nine);
		\draw[->] (nine) to (ten);
		\draw[->] (ten) to (eleven);
	\end{tikzpicture}
	\caption{Depiction of the coagent network for the hierarchical option-critic architecture. }
	\label{fig:coagent-hiararcy-hocpg}
\end{figure*}

In Figure \ref{fig:coagent-hiararcy-hocpg} we depict an abstract view of the co-agent network for a hierarchy of options. Each $\beta^\ell$ will receive an input indicating whether it has to decide to terminate or not, otherwise it will act as a pass-through policy outputting NULL. After we have passed the upward phase of execution (deciding which options to terminate), we begin the downward phase of selecting new options at levels where the previous option terminated. For the downward phase we maintain a vector called the termination vector $T$ of zeros (termination), a single one at some level of abstraction $\ell$ (no termination), and the rest of the elements are filled with NULL. This vector along with $o^{1:N-1}$ is passed through each $\pi^\ell$. The vector indicates at which $\ell$ we have to start picking new option $o_{t+1}^\ell$ and replacing the old one $o_{t}^\ell$. 
Finally, a primitive action is chosen with $\pi^N$ after all options policies have been processed.

Alternatively, we can actually write the local policy gradient theorem for all of $\pi^{N-1} \circ ... \circ \pi^{1} \circ \beta^{1} \circ ... \circ \beta^{N-1} = \kappa^{1:N-1}$ and $\pi^N = \kappa^N$. The objectives of these terms can be added together in order to recover the full shared parameter policy gradient theorem. This allows us to decompose the gradient with respect to the total objective function as follows by subbing into equation \ref{eq:capg}: 

\begin{equation}
\begin{split}
 \frac{\partial J_{1:N} }{\partial \bm{\theta}} = \frac{\partial J_N }{\partial \bm{\theta}} + \frac{\partial J_{1:N-1} }{\partial \bm{\theta}} \\ 
 \frac{\partial J_N }{\partial \bm{\theta}} = \sum_{s, o^{1:N-1}, s'} \mu_\Omega (s, o^{1:N-1}, s'| s_0, o_0^{1:N-1}) \bigg(\\ \sum_a \frac{\partial \pi^N(a|s,o^{1:N-1})}{\partial \bm{\theta}} Q_U(s, o^{1:N-1}, a) \bigg) \\
 \frac{\partial J_{1:N-1} }{\partial \bm{\theta}} = \sum_{s, o^{1:N-1}, s'} \mu_\Omega (s, o^{1:N-1}, s'| s_0, o_0^{1:N-1}) \sum_{o'^{1:N-1}} \bigg( \\ \frac{\partial \kappa^{1:N-1}(o'^{1:N-1}|s',o^{1:N-1})}{\partial \bm{\theta}} Q_\Omega(s', o'^{1:N-1}) \bigg)
\end{split}
\end{equation}

Interestingly, $\kappa^{1:N-1}$ was defined earlier as $P_{\pi,\beta}(o'^{1:N-1}|s',o^{1:N-1})$, which is the probability while at the next state and terminating the options for the current state that the agent arrives at a particular set of next option selections. Subbing this notation in we find that: 

\begin{equation}
\begin{split}
\frac{\partial J_{1:N-1} }{\partial \bm{\theta}} = \sum_{s, o^{1:N-1}, s'} \mu_\Omega (s, o^{1:N-1}, s'| s_0, o_0^{1:N-1}) \bigg( \\ \sum_{o'^{1:N-1}} \frac{\partial P_{\pi,\beta}(o'^{1:N-1}|s',o^{1:N-1})}{\partial \bm{\theta}} Q_\Omega(s', o'^{1:N-1}) \bigg )
\end{split}
\end{equation}

We will now show that $\frac{\partial J_{1:N} }{\partial \bm{\theta}}$ is exactly the same as the objective learned by Theorem 1. To explain why this is we start by providing useful definitions of $Q_\Omega$ and $Q_U$:

\begin{equation}
\begin{split}
Q_\Omega(s,o^{1:N-1}) = \sum_{a} \pi(a|s,o^{1:N-1}) Q_\Omega(s,o^{1:N-1},a)
\end{split}
\end{equation}

\begin{equation}
\begin{split}
Q_U(s',o^{1:N-1}) = \\ \sum_{o'^{1:N-1}} P_{\pi,\beta}(o'^{1:N-1}|s',o^{1:N-1}) Q_\Omega(s',o'^{1:N-1})
\end{split}
\end{equation}

Next we take the derivative of $Q_\Omega$ and then note how the right hand term can be expanded: 

\begin{equation}
\begin{split}
 \frac{\partial Q_\Omega(s,o^{1:N-1})}{\partial \bm{\theta}} = \sum_{a} \frac{\partial \pi^N(a|s,o^{1:N-1})}{\partial \bm{\theta}} Q_\Omega(s,o^{1:N-1},a) \\ + \sum_{a} \pi^N(a|s,o^{1:N-1}) \frac{\partial Q_\Omega(s,o^{1:N-1},a)}{\partial \bm{\theta}} \\
 \sum_{a} \pi^N(a|s,o^{1:N-1}) \frac{\partial Q_\Omega(s,o^{1:N-1},a)}{\partial \bm{\theta}} = \\ \sum_{a} \pi^N(a|s,o^{1:N-1}) \sum_{s'} P(s'|s,a) \frac{\partial Q_U(s',o^{1:N-1})}{\partial \bm{\theta}}
\end{split}
\end{equation}

To expand this expression further we use the product rule to take the derivative with respect to $Q_U$:

\begin{equation}
\begin{split}
\frac{\partial Q_U(s',o^{1:N-1})}{\partial \bm{\theta}} = \\ \sum_{o'^{1:N-1}} \frac{\partial P_{\pi,\beta}(o'^{1:N-1}|s',o^{1:N-1})}{\partial \bm{\theta}} Q_\Omega(s',o'^{1:N-1}) \\ + \sum_{o'^{1:N-1}} P_{\pi,\beta}(o'^{1:N-1}|s',o^{1:N-1}) \frac{\partial Q_\Omega(s',o'^{1:N-1})}{\partial \bm{\theta}}
\end{split}
\end{equation}

Subbing this back into the right hand side of our original expression yields:

\begin{equation}
\begin{split}
\sum_{a} \pi^N(a|s,o^{1:N-1}) \frac{\partial Q_\Omega(s,o^{1:N-1},a)}{\partial \bm{\theta}} = \\ \sum_{a} \pi^N(a|s,o^{1:N-1}) \sum_{s'} P(s'|s,a) \bigg[ \\ \sum_{o'^{1:N-1}} \frac{\partial P_{\pi,\beta}(o'^{1:N-1}|s',o^{1:N-1})}{\partial \bm{\theta}} Q_\Omega(s',o'^{1:N-1}) \\ + \sum_{o'^{1:N-1}} P_{\pi,\beta}(o'^{1:N-1}|s',o^{1:N-1}) \frac{\partial Q_\Omega(s',o'^{1:N-1})}{\partial \bm{\theta}} \bigg]
\end{split}
\end{equation}

This implies the following expression for the full derivative: 

\begin{equation}
\begin{split}
 \frac{\partial Q_\Omega(s,o^{1:N-1})}{\partial \bm{\theta}} = \sum_{a} \frac{\partial \pi^N(a|s,o^{1:N-1})}{\partial \bm{\theta}} Q_\Omega(s,o^{1:N-1},a) \\ + \sum_{a} \pi^N(a|s,o^{1:N-1}) \sum_{s'} P(s'|s,a) \bigg[ \\ \sum_{o'^{1:N-1}} \frac{\partial P_{\pi,\beta}(o'^{1:N-1}|s',o^{1:N-1})}{\partial \bm{\theta}} Q_\Omega(s',o'^{1:N-1}) \\+ \sum_{o'^{1:N-1}} P_{\pi,\beta}(o'^{1:N-1}|s',o^{1:N-1}) \frac{\partial Q_\Omega(s',o'^{1:N-1})}{\partial \bm{\theta}} \bigg]
\end{split}
\end{equation}

Following the analysis of the recursive augmented process earlier in the appendix, we can simplify the previous expression: 

\begin{equation}
\begin{split}
\frac{\partial Q_\Omega(s,o^{1:N-1})}{\partial \bm{\theta}} =\mu_\Omega (s, o^{1:N-1}, s'| s_0, o_0^{1:N-1}) \bigg( \\ \sum_{a} \frac{\partial \pi^N(a|s,o^{1:N-1})}{\partial \bm{\theta}} Q_\Omega(s,o^{1:N-1},a) \\ + \sum_{o'^{1:N-1}} \frac{\partial P_{\pi,\beta}(o'^{1:N-1}|s',o^{1:N-1})}{\partial \bm{\theta}} Q_\Omega(s',o'^{1:N-1}) \bigg) \\ 
\frac{\partial Q_\Omega(s,o^{1:N-1})}{\partial \bm{\theta}} = \frac{\partial J_{1:N} }{\partial \bm{\theta}}
\end{split}
\end{equation}

Now we can see that the co-agents policy gradient theorem objective is indeed $\frac{\partial Q_\Omega(s,o^{1:N-1})}{\partial \bm{\theta}}$, which is precisely the starting point of the derivation in earlier in the appendix. As such, the co-agent policy gradient theorem also yields the following update:

\begin{equation}
\begin{split}
\frac{\partial J_{1:N} }{\partial \bm{\theta}} = \frac{\partial Q_\Omega(s,o^{1:N-1})}{\partial \bm{\theta}} = \\ \sum_{s,o^{1:N-1},s'}  \mu_\Omega(s,o^{1:N-1},s'|s_0,o_0^{1:N-1})  \bigg( \\ \sum_{a} \frac{\partial \pi(a|s,o^{1:N-1})}{\partial \bm{\theta}} Q_U(s,o^{1:N-1},a) \\ + \gamma \sum_{o'^{1:N-1}} \sum_{\ell=1}^{N-1} \frac{\partial \pi^\ell(o'^\ell|s',o'^{1:\ell-1})}{\partial \bm{\theta}} Q_\Omega(s',o'^{1:\ell}) \bigg[ \\ \prod_{k=N-1}^{\ell} \beta^k(s',o^{1:k}) \bigg] \\ - \gamma \sum_{\ell=1}^{N-1} \frac{\partial \beta^\ell(s',o^{1:\ell})}{\partial \bm{\theta}} A_\Omega(s',o^{1:\ell}) \prod_{k=N-1}^{\ell+1} \beta^k(s',o^{1:k}) \bigg).
\end{split}
\end{equation}

\section{Appendix D: Algorithm Implementation Details}

\subsection{Algorithm Details}

In algorithm \ref{AtariAlg} we present our modified version of Asynchronous Advantage Option-Critic (A2OC) \citep{Deliberation}, which is very similar the original version, but with a single more general update for the full system. We follow a common convention \citep{A3C} and set $\alpha_v$ the relative learning rate of the value function equal to 0.5 in our experiments. $\eta$ refers to the regularizer on the termination function update. The term $A_\Omega(s_{t+1},o_t)$ is approximated by leveraging the fact that $A_\Omega(s_{t+1},o_t)=Q_\Omega(s_{t+1},o_t)-V_\Omega(s_{t+1})=Q_\Omega(s_{t+1},o_t)-max_{o_t} Q_\Omega(s_{t+1},o_t)$, allowing us to maintain only a function approximation of $Q_\Omega(s,o)$ explicitly. In order to convert the sums in Lemma 1 into an expectation we take the log of the policies to account for the sums over $a$ and $o'$. 

Additionally, in algorithm \ref{AtariHOCAlg} we also present a modified version of Asynchronous Advantage Hierarchical Option-Critic (A2HOC) \citep{HOC} with our new update rule from Theorem 1. Following past work, we leverage the fact that $A_\Omega(s',o^{1:\ell}) = Q_\Omega(s',o^{1:\ell}) - V_\Omega(s') [\prod_{j=\ell-1}^{1} \beta^j(s',o^{1:j})]  - \sum_{i=1}^{\ell - 1}(1-\beta^i(s',o^{1:i})) Q_\Omega(s',o^{1:i}) [\prod_{k=i+1}^{\ell - 1} \beta^k(s',o^{1:k})]$ to approximate $A_\Omega(s',o^{1:\ell})$. In our experiments we chose to represent each $Q_\Omega(s',o^{1:i})$ for $1<i<N-1$ with different function approximators (i.e. a single fully connected layer) that takes in the state representation from the CNN and LSTM with an output predicting the value of each option. When necessary, we approximated $V_\Omega(s_{t})$ by leveraging the fact that $V_\Omega(s_{t})=max_{o_t} Q_\Omega(s_{t},o_t)$. As before when converting to an expectation, taking the log of the policy addresses the sum over $a$. We address the sum over $o^{1:N-1}$ by taking the log of the current policy at level $\ell$ which accounts for the sum over $o^{\ell}$, the sum over higher level options are addressed by $P_{\pi,\beta}$ and the term has no expected dependence on the lower level options. 

\begin{algorithm}
  \caption{A2OC with Option-Critic Policy Gradients }\label{AtariAlg}
  \begin{algorithmic}
  \Procedure{Learn}{$env,N,\alpha,\alpha_v,\gamma,\pi,\beta,\eta$} 
  \State initialize global counter $T \gets 1$
  \State initialize thread counter $t \gets 1$
  \Repeat
  \State $t_{start} = t$
  \State $s_t \gets s_0$
  \State // reset gradients
  \State $d\bm{\theta} \gets 0$
  \State // select option for initial state
  \State $o_t \gets \pi_\Omega(s_t)$
  \Repeat
  \State // take an action
  \State $a_t \gets \pi(s_t,o_t)$
  \State // step through the environment
  \State $s_{t+1},r_t \gets env.step(a_t)$
  \State // check if the option has terminated
  \State \textbf{if} $\beta(s_{t+1},o_t) = 1$
  \State\hspace{\algorithmicindent}  // select a new option
  \State\hspace{\algorithmicindent} $o_{t+1} \gets \pi_\Omega(s_{t+1})$
  \State $t \gets t + 1$
  \State $T \gets T + 1$
  \Until{episode ends or $t - t_{start} == t_{max}$ or $(t - t_{start} > t_{min}$)          }
  \State $G = V(s_t)$ 
    \For{\texttt{$k = t-1,...,t_{start}$}}
  \State // accumulate thread specific gradients
  \State $G \gets r_k + \gamma G$ 
  \State // update the approximate action-value function                  
  \State $dw \gets dw + \alpha_w \frac{\partial (G - Q(s_t, o_t))^2}{\partial w}$
  \State // unified policy gradient update
  \State $d\bm{\theta} \gets d\bm{\theta} + \alpha \bigg( \frac{\partial log\pi(a_t|s_t,o_t)}{\partial \bm{\theta}} (G - Q_\Omega(s_t,o_t)) + \gamma \beta(s_{t+1},o_t) \frac{\partial log\pi_\Omega(o_{t+1}|s_{t+1})}{\partial \bm{\theta}} (G - V_\Omega(s_{t+1}))$
  $- \gamma\frac{\partial \beta(s_{t+1},o_t)}{\partial \bm{\theta}}(A_\Omega(s_{t+1},o_t)+\eta) + \alpha_v \frac{\partial (G - Q(s_t, o_t))^2}{\partial \bm{\theta}} \bigg)$
  \EndFor
  \State Update global parameters with thread gradients
  \Until{$T > T_{max}$}
  \EndProcedure
  \end{algorithmic}
\end{algorithm}

\begin{algorithm}
  \caption{A2HOC with Hierarchical Option-Critic Policy Gradients }\label{AtariHOCAlg}
  \begin{algorithmic}
  \Procedure{Learn}{$env,N,\alpha,\gamma,\pi,\beta,\eta,T_{max},t_{min},t_{max}$}
  \State initialize global counter $T \gets 1$
  \State initialize thread counter $t \gets 1$
  \Repeat
  \State $t_{start} = t$
  \State $s_t \gets s_0$
  \State // reset gradients
  \State $dw \gets 0$ 
  \State $d\theta \gets 0$
  \State $d\phi \gets 0$
  \State // select options for initial state
  \For{\texttt{$j = 1,...,N-1$}}
  \State $o_t^{j} \gets \pi^{j}(s_t,o_t^{1:j-1})$
  \EndFor
  \Repeat
  \State // take an action
  \State $a_t \gets \pi^N(s_t,o_t^{1:N-1})$
  \State // step through the environment
  \State $s_{t+1},r_t \gets env.step(a_t)$
  \State // check which options have terminated 
  \State $o_t^{1:N-1} \gets newOptions(s_{t+1},o_{t-1}^{1:N-1},\pi,\beta, N)$
  \State $t \gets t + 1$
  \State $T \gets T + 1$
  \Until{episode ends or $t - t_{start} == t_{max}$ or $(t - t_{start} > t_{min}$)          }
  \State $G = V(s_t)$ 
  \For{\texttt{$k = t-1,...,t_{start}$}}
  \State // accumulate thread specific gradients
  \State $G \gets r_k + \gamma G$ 
  \State // unified policy gradient update
  \State $d\bm{\theta} \gets d\bm{\theta} + \alpha \bigg( \frac{\partial log\pi(a_t|s_t,o_t^{1:N-1})}{\partial \bm{\theta}} (G - Q_\Omega(s_t,o_t^{1:N-1})) + \gamma \sum_{\ell=1}^{N-1} \prod_{k=N-1}^{\ell} \beta^k(s',o^{1:k}) \frac{\partial log\pi^\ell(o'^\ell|s',o'^{1:\ell-1})}{\partial \bm{\theta}} (G - Q_\Omega(s',o'^{1:\ell-1})) - \gamma \sum_{\ell=1}^{N-1} \frac{\partial \beta^\ell(s',o^{1:\ell})}{\partial \bm{\theta}} (A_\Omega(s',o^{1:\ell})+\eta) \prod_{k=N-1}^{\ell+1} \beta^k(s',o^{1:k}) + \alpha_v \sum_{\ell=1}^{N-1} \frac{\partial (G - Q(s_t, o_t^{1:\ell}))^2}{\partial \bm{\theta}} \bigg)$
  \EndFor
  \State update global parameters with thread gradients
  \Until{$T > T_{max}$}
  \EndProcedure
  \Procedure{newOptions}{$s,o^{1:k},\pi,\beta, k$} 
    
    \State \textbf{if} $\beta^k(s,o^{1:k}) = 1$
    \State\hspace{\algorithmicindent} \textbf{if} $k-1 = 1$
   \State\hspace{\algorithmicindent}\hspace{\algorithmicindent} $o^1 \gets \pi^{1}(s)$
    \State\hspace{\algorithmicindent} \textbf{else}
    \State\hspace{\algorithmicindent}\hspace{\algorithmicindent} $o^{1:k-1} \gets newOptions(s,o^{1:k-1},\pi,\beta,k-1)$
    \State\hspace{\algorithmicindent} $o^k \gets \pi^{k-1}(s,o^{1:k-1})$

    \State \textbf{return} $o^{1:k}$
    \EndProcedure
  \end{algorithmic}
\end{algorithm}

\subsection{Computational Complexity Analysis} \label{AppendixComplexity} 

We will now analyze the computational complexity of the option-critic policy gradient theorem (Lemma 1) update rule. We begin by making some assumptions that simplify our analysis and elucidate the practical implications of implementing the option-critic policy gradient theorem. We will then demonstrate that the complexity of Algorithms 1 and 2 are quite similar to past work on option-critic learning and empirically evaluate run time characteristics to validate that this is indeed true in practice as well. 

\subsubsection{Notation and Assumptions} 

In this section we are primarily interested in analyzing the amount of computation per step $\mathscr{C}$ needed to update an agent leveraging the option-critic architecture following Lemma 1 or Theorem 1. We will denote the computation needed for Lemma 1 as $\mathscr{C}_{OCPG}$ and the computation needed for the general hierarchical version in Theorem 1 as $\mathscr{C}_{HOCPG}$. We will denote the forward propagation computation of a neural network as $\mathscr{F}$ while using the subscript notation to refer to the forward propogation of a particular architecture component in the set of policies $\mathscr{F}_{\boldsymbol{\pi}}$, the set of termination functions $\mathscr{F}_{\boldsymbol{\beta}}$, and the set of Q functions $\mathscr{F}_{\boldsymbol{Q}}$. Likewise, we will denote the backpropagation computation of a neural network which is invoked to compute the derivative of the selected action for each architecture component with respect to its own parameters as $\mathscr{B}$. We again use the subscript notation to refer to the backpropagation of a particular architecture component in the set of policies $\mathscr{B}_{\boldsymbol{\pi}}$, the set of termination functions $\mathscr{B}_{\boldsymbol{\beta}}$, and the set of Q functions $\mathscr{B}_{\boldsymbol{Q}}$. We now make two assumptions that are reasonable in practice in order to simplify our analysis of the complexity for each model: 

\textbf{Assumption 1:} We will assume that the network used to extract features for each architecture component in $\boldsymbol{\pi}$, $\boldsymbol{\beta}$, $\boldsymbol{Q}$ is of very similar size and complexity. While there could be benefits moving forward in using architectures for which this assumption does not hold, this has always been the case for past work on option learning as customarily the same network or a parallel network is used to extract a final hidden representation $h$ for each architecture component. 

\textbf{Assumption 2:} We will assume that each architecture component in $\boldsymbol{\pi}$, $\boldsymbol{\beta}$, $\boldsymbol{Q}$ contains a fully connected final output layer and that the parameters contained within it are much less than the parameters contained in the network that produces $h$. Again, this is typically true in practice for work on option-critic learning with deep function approximation. 

\textbf{Implications of Assumptions 1 and 2:} With these two assumptions in mind, we can see that differences in the number of units in the final classification layer should have minimal effect on computation. As such, we expect both the forward propagation and backpropagation computations to be of very similar magnitude across architecture components so we can approximately say that $\mathscr{F}_{\boldsymbol{\pi}}=\mathscr{F}_{\boldsymbol{\beta}}=\mathscr{F}_{\boldsymbol{Q}}=\mathscr{F}$ and that $\mathscr{B}_{\boldsymbol{\pi}}=\mathscr{B}_{\boldsymbol{\beta}}=\mathscr{B}_{\boldsymbol{Q}}=\mathscr{B}$. 

\textbf{Assumption 3:} Next, we assume the the computation of backpropagation gradients is much bigger than computation of forward propagation $\mathscr{B}>>\mathscr{F}$. This assumption is invariably true in practice today using modern neural network software packages. 

\textbf{Assumption 4:} Finally, we assume that the computational complexity of multiplying by a constant value (i.e. $\gamma$) is negligible relative to the computational cost of forward propagating through the network $\mathscr{F} >> \mathscr{C}_{\gamma}$. This assumption is once again invariably true in practice when using modern neural networks with highly non-trivial size. 

\subsubsection{Analysis of Complexity} 

First let us consider the terms included in Lemma 1 to understand $\mathscr{C}_{OCPG}$. The first term includes the forward propagation of $Q_U$ and the backpropagation of $\pi$. Following assumption 4, we find that the second term then includes the forward propagation of $Q_\Omega$ and $\beta$ as well as the backpropagation of $\pi_{\Omega}$. The third term then includes the forward propagation of $A_\Omega$ and backpropagation of $\beta$.

\begin{equation} \label{COCPG}
\begin{aligned}
\mathscr{C}_{OCPG} &\approx \mathscr{F}_{Q_U} + \mathscr{B}_{\pi} + \mathscr{F}_{Q_\Omega} + \mathscr{F}_{\beta} + \mathscr{B}_{\pi_{\Omega}} + \mathscr{F}_{A_\Omega} + \mathscr{B}_{\beta} \\ 
&\approx \mathscr{F} + \mathscr{B} + \mathscr{F} + \mathscr{F} + \mathscr{B} + \mathscr{F} + \mathscr{B} = 3\mathscr{B} + 4\mathscr{F} \\
&\approx 3\mathscr{B}
\end{aligned}
\end{equation}

As seen in equation \ref{COCPG}, this expression could be further simplified to $3\mathscr{B} + 4\mathscr{F}$ by leveraging assumptions 1 and 2. Finally, we consider assumption 3 and realize the forward propagation time will hardly influence the total computation time of the update rule to discover that $\mathscr{C}_{OCPG} \approx 3\mathscr{B}$. 

Now we will consider the terms included in Theorem 1 to understand $\mathscr{C}_{HOCPG}$. We consider the computation involved for arbitrary level of abstraction $N>3$. The expression becomes a bit more complex, including summations across the hierarchical levels of abstraction and summations over lower level options that need to terminate before an architectural component is used: 

\begin{equation}  \label{CHOCPG}
\begin{aligned}
\mathscr{C}_{HOCPG} &\approx \mathscr{F}_{Q_U^N} + \mathscr{B}_{\pi^N} + \sum_{\ell=1}^{N-1} \bigg[ \mathscr{F}_{Q_\Omega^\ell} + \mathscr{B}_{\pi^\ell} + \sum_{k=N-1}^\ell \mathscr{F}_{\beta^k} \bigg] \\ &+  \sum_{\ell=1}^{N-1} \bigg[ \mathscr{F}_{A_\Omega^\ell} + \mathscr{B}_{\beta^\ell} \sum_{k=N-1}^{\ell+1} \mathscr{F}_{\beta^k} \bigg] \\ 
&\approx \mathscr{F} + \mathscr{B} + (N-1) \mathscr{F} + (N-1) \mathscr{B} \\ &+ \frac{1}{2} (N-2) (N-1) \mathscr{F} + (N-1) \mathscr{F} + (N-1) \mathscr{B} \\ &+ \frac{1}{2} (N-2) (N-3) \mathscr{F}\\
&= (2N-1) \mathscr{F} + (2N-1) \mathscr{B}  + \frac{1}{2} (N^2 - 3N + 2) \mathscr{F} \\ &+ \frac{1}{2} (N^2 - 5N + 6) \mathscr{F} \\
&= (2N-1) \mathscr{B} + (N^2 - 2N + 3) \mathscr{F} \\
&\approx (2N-1) \mathscr{B} 
\end{aligned}
\end{equation}

As seen in equation \ref{CHOCPG}, this expression could be further simplified to $(2N-1) \mathscr{B} + (N^2 - 2N + 3) \mathscr{F}$ by leveraging assumptions 1 and 2. We must now considered a slightly stronger assumption than assumption 3 i.e. $\mathscr{B} >> N \mathscr{F}$. This has always been true in past work on building deep hierarchies of options with neural networks that have focused on relatively small values of $N$. Unfortunately, there are multiple computational barriers associated with achieving $N \rightarrow \infty$, so this assumption is reasonable in practice. Once again, forward propagation time will hardly influence the total computation time of the update rule and we discover that $\mathscr{C}_{HOCPG} \approx (2N-1) \mathscr{B}$. 

\subsubsection{Comparing the Complexity of Other Approaches} 

We would now like to consider the computation time of the update rule for the standard option-critic architecture \citep{OC} i.e. $\mathscr{C}_{OC}$. The terms are quite similar with the exception of not including some constant multiplications by $\gamma$, which is assumed to be of negligible in assumption 4, and not including the forward propagation of $\beta$ during the update of the policy over options. 

\begin{equation} 
\begin{aligned}
\mathscr{C}_{OC} &\approx \mathscr{F}_{Q_U} + \mathscr{B}_{\pi} + \mathscr{F}_{Q_\Omega} + \mathscr{B}_{\pi_{\Omega}} + \mathscr{F}_{A_\Omega} + \mathscr{B}_{\beta} \\ 
&\approx \mathscr{F} + \mathscr{B} + \mathscr{F} + \mathscr{B} + \mathscr{F} + \mathscr{B} = 3\mathscr{B} + 3\mathscr{F} \\
&\approx 3\mathscr{B} \approx \mathscr{C}_{OCPG}
\end{aligned}
\end{equation}

Again, this expression can be simplified using assumptions 1 and 2 to $3\mathscr{B} + 3\mathscr{F}$. Finally, we consider assumption 3 to discover that $\mathscr{C}_{OC} \approx 3\mathscr{B} \approx \mathscr{C}_{OCPG}$. In section \ref{EmpiricalComplexity} we back up this theoretical analysis with empirical run time performance statistics, demonstrating very similar computational requirements to the standard option-critic update for the option-critic policy gradient theorem update. 

We would finally like to consider the computation time of the update rule for the standard hierarchical option-critic architecture \citep{HOC} i.e. $\mathscr{C}_{HOC}$. The terms are quite similar to equation \ref{CHOCPG} while not including some constant multiplications by $\gamma$, which is assumed to be of negligible in assumption 4, and not including the forward propagation of the termination functions $\beta$ for lower level options.  

\begin{equation} 
\begin{aligned}
\mathscr{C}_{HOC} &\approx \mathscr{F}_{Q_U^N} + \mathscr{B}_{\pi^N} \\ &+ \sum_{\ell=1}^{N-1} \bigg[ \mathscr{F}_{Q_\Omega^\ell} + \mathscr{B}_{\pi^\ell} \bigg] +  \sum_{\ell=1}^{N-1} \bigg[ \mathscr{F}_{A_\Omega^\ell} + \mathscr{B}_{\beta^\ell} \bigg] \\ 
&\approx \mathscr{F} + \mathscr{B} + (N-1) \mathscr{F} + (N-1) \mathscr{B} \\ &+ (N-1) \mathscr{F} + (N-1) \mathscr{B} \\
&= (2N-1) \mathscr{F} + (2N-1) \mathscr{B} \\
&\approx (2N-1) \mathscr{B} \approx \mathscr{C}_{HOCPG}
\end{aligned}
\end{equation}

Leveraging assumptions 1 and 2, this expression can be simplified to to $(2N-1) \mathscr{F} + (2N-1) \mathscr{B}$. Considering assumption 3, we can see that $\mathscr{C}_{HOC} \approx (2N-1) \mathscr{B} \approx \mathscr{C}_{HOCPG}$.

\subsubsection{Computation Comparisons in Practice} \label{EmpiricalComplexity}

In order to get an empirical sense of the degree to which the computational complexity of the option-critic policy gradient theorem is similar to the standard option-critic update rule in practice, we monitor the speed of progression through the three Atari games used in our experiments with a regularization schedule. For each game, we measure 4 random runs of both OC and OCPG where we allow the agent to train for 1 day across 16 threads. We then report the average and standard deviation of the number of steps in the game taken per minute per thread (Table \ref{Computation}). We focus on progress through 15 threads for each random run (60 threads in total). This is because one thread was used to save the model as well and we do not include it in the average as it consistently achieved less progression through the game than the other 15 threads. We see some degree of variation across games as a result of game dependent simulator characteristics. However, we clearly see very comparable run time characteristics for both OC and OCPG across games and runs. 

\begin{table}
\centering
\begin{tabular}{|c|c|c|}
\hline
Game & OC & OCPG  \\ \hline
\normalsize Alien & \normalsize  1,243.5	\tiny{$\pm$ 137.9} \normalsize & \normalsize  1,315.6 \tiny{$\pm$ 33.7} \normalsize \\ \hline
\normalsize Amidar & \normalsize  1,290.1	\tiny{$\pm$ 45.1} \normalsize & \normalsize  1,212.3 \tiny{$\pm$ 163.4} \normalsize \\ \hline
\normalsize Tutankham & \normalsize 1,235.7	\tiny{$\pm$ 153.5} \normalsize & \normalsize  1,240.1 \tiny{$\pm$ 142.6} \normalsize \\ \hline
\end{tabular}
\caption{ Average steps per minute per thread for 3 Atari games across 60 threads of parallel training on CPU. } 
\label{Computation}
\end{table}

\section{Appendix E: Additional Experiment Details }

\subsection{Architecture Details}

We leverage the popular Open AI Gym interface to the games and use the default setting for each game. We use standard pre-processing of the Atari environment input and reward signal to allow for a single learning rate across games \citep{DQN}. Our architecture follows \citet{DQN} as well consisting of a feature extractor common across all components of the architecture with 4 convolutional layers each followed by a max pooling and ReLU layer fed into an LSTM as in \citep{A3C}. As in past work, each policy, termination function, and value function has its own fully connected layer on top of this base feature extractor. The first convolutional layer has a 5x5 kernel with 32 filters, stride of 1, and padding of 2. The second layer is the same as the first, but with a padding of 1. The third layer has 64 filters, a 4x4 kernel, stride of 1, and padding of 1. The fourth layer is the same as the third, but with a 3x3 kernel. All max pooling layers are 2x2. The LSTM has 512 hidden units. 

\subsection{Computing Infrastructure}

The experiments in this paper were conducted by leveraging a cluster of x86 Intel compute nodes in which jobs were run for multiple days on 16 parallel CPU cores. 

\subsection{Environments for Our Experiments} 

For our Atari experiments, we leverage the standard "v0" Open AI Gym version of these environments. For documentation on how to get started with these environments see \url{https://gym.openai.com/envs/#atari}. 

\subsection{Instability Issues During Training} 

In our experiments for both OCPG and OC, we see some degree of instability during training. This is an systematic issue for on-policy learning methods in general and has been overcome in the Deep RL literature primarily by leveraging off-policy learning with experience replay as in \citet{DQN}. While this instability issue is somewhat mitigated by maintaining many parallel threads of learning, all of our agents are still prone to catastrophic forgetting \citep{CF} without stabilizing learning by re-learning past experiences while learning the current experience. While naive experience replay may suffer from issues related to the differences in the policy over time and issues with scaling storage, these problems have been addressed in the literature by leveraging importance sampling \citep{is1,is2,is3,importancesampling} and scalable storage mechanisms \citep{Robins95,LwF,FC,GenerativeDist,Shin,Recollections} respectively. Our policy gradient theorems proposed in this work should allow for easy integration with any of these approaches when suitable for increased stability during learning. 

Modularity in representations has also been a very effective tool for learning multiple (possibly interfering) skills in a neural network \citep{ebengio,HLS,cross,PNN,largeneuralnets,PathNets,RoutingNetworks,SLUICE,diversity,NAACL19} that directly addresses catastrophic forgetting, increasing representation stability during learning. See \citep{challenges} for a survey of challenges in learning this kind of representation end to end. As explained in the main text, our new OCPG policy gradient theorem allows for very dynamic assumptions about weight sharing during learning and enables direct optimization of the correct policy gradient for these modular models. While we have not explored this extension in our experiments, OCPG should theoretically make way for models that are increasingly robust to stability issues during learning by directly considering more dynamic weight sharing schemes.   

\subsection{Analysis of Learned Options} 

\begin{figure*}
    \centering
    \includegraphics[width=1.0\textwidth]{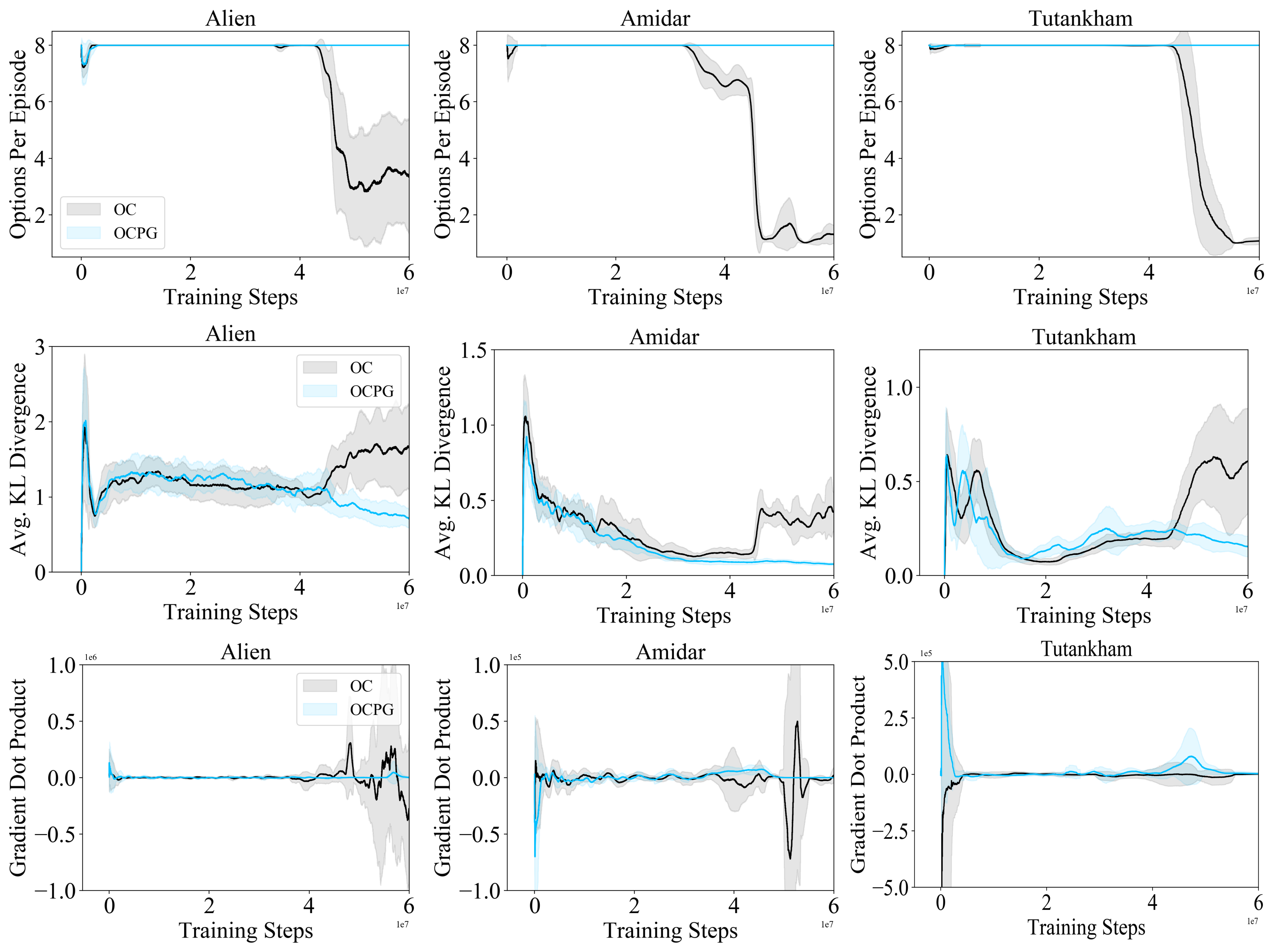}
    \caption{Further analysis of the options that are learned during our experiments with a regularization schedule. The first row of charts reports the number of distinct options used per episode during training. The second row reports the average pairwise KL divergence between the policy for each option over time. Finally, the third row reports the average dot product of the gradient for learning $\pi_\Omega$ with a reservoir of past gradients for $\pi_\Omega$ during learning. }
    \label{figure:options}
\end{figure*}

To provide further clarity about our experiments with a regularization schedule in the main text, we report additional interesting details about these experiments in Figure \ref{figure:options}. All results in Figure \ref{figure:options} are reported leveraging the testing thread over the course of learning.  In the first row of Figure \ref{figure:options} we report the average number of distinct options used per episode over the course of training. This analysis reveals a very clear difference between the behavior OCPG and OC. Early in training, after random initialization, OCPG and OC both leverage all 8 available options during a typical episode. However, once the regularization becomes high later in training, we see that OC starts only using one or a few options per episode. Meanwhile, OCPG is able to still use all of the options even as the regularization promotes options that are longer.

In order to understand more about why this is the case, in the second row of Figure \ref{figure:options} we provide the average pairwise KL divergence of all options with every other option at the states encountered by the agents over the course of training. In all cases, we again see a consistent pattern later in training when the regularization is high. When options get longer, OC learns options that become increasingly differentiated from each other. On the other hand, OCPG is better able to exploit commonalities and weight sharing between options even when they become longer. For OCPG, we actually see the exact opposite behavior as the options even seem to become slightly more similar as they become longer. 

Finally, to provide additional clarity, in the final row of Figure \ref{figure:options} we follow \citep{MER} and try to get a sense of the transfer and interference behavior between gradients over the course of learning. In particular, we focus on gradients for the policy over options ($\pi_\Omega$) as this part of the gradient accounts for the key difference between the OC and OCPG update rules. During learning, we compute gradients for $\pi_\Omega$ after each episode and compare it to 5 random gradients sampled from a buffer of 20 historical gradients that is maintained leveraging reservoir sampling. Following \citep{MER}, we can say that when the gradient dot product is positive, this results in transfer between experiences and when the gradient dot product is negative this leads to interference between experiences. As we would expect, the instability in performance that OC experiences later in training when options are longer is directly related to a more negative distribution of gradient dot products between current and past experiences. Additionally, we can also notice that OCPG often experiences a thinner (and more stable) distribution of gradient dot products later in training. Both of these advantages are intuitive based on the different forms of the OCPG and OC update rules. Because OCPG gates the gradients of the policy over options by the probability that the underlying options actually terminate, we would expect a thinner distribution of gradient dot products overall. Moreover, gradients for the policy over options when it is unlikely to be used are more likely to create interference with past learning as these experiences are largely irrelevant. 

\end{document}